\documentclass[letterpaper]{article}
\usepackage{preprint}
\usepackage[hyphens]{url}
\usepackage{graphicx}
\usepackage{natbib}
\usepackage{caption}
\usepackage{amsmath,amssymb}
\usepackage{amsthm}
\usepackage{booktabs}
\usepackage{colortbl}
\usepackage{tabularx}
\usepackage{longtable}
\usepackage{multirow}
\usepackage{float}
\usepackage[hidelinks]{hyperref}

\title{Prompt-Anchored Residual Adaptation for Biomedical Vision-Language
Models}

\author{
    Jingxuan Kang\textsuperscript{\rm 1},
    Qianying Yue\textsuperscript{\rm 2},
    Che Liu\textsuperscript{\rm 1},
    Chen Qin\textsuperscript{\rm 1}\corresponding
}
\affiliations{
    \textsuperscript{\rm 1}Imperial College London\\
    \textsuperscript{\rm 2}The Chinese University of Hong Kong\\
    j.kang26@imperial.ac.uk, c.qin15@imperial.ac.uk
}

\begin{document}

\maketitle

\begin{abstract}
Pretrained biomedical vision--language models achieve strong zero-shot performance in biomedical image classification. However, downstream biomedical classification often depends on subtle visual differences between classes that may not be fully captured by pretrained representations. Few-shot adaptation addresses this mismatch by optimizing a task-specific predictor on a small labeled support set. Because the selected examples capture only part of the visual variation within the target classes, the adapted predictions can depend strongly on their composition. We propose \textbf{Prompt-Anchored Residual Adaptation (PARA)}, which retains the frozen prompt prediction as a support-invariant semantic anchor and incorporates a visual prediction learned from the support set through an anchor-relative residual. The residual step is computed in a closed form from frozen support embeddings using anchor discrepancy and support agreement. Support-set dependence also limits evaluation: comparisons are fair within a shared draw but remain conditional on its composition. To obtain more reliable comparisons, we introduce a repeated-support protocol that separates support-selection variation from optimization randomness and reports both average and worst-20\% performance. PARA achieves state-of-the-art performance in both few-shot classification and base-to-novel generalization.
\end{abstract}

\section{Introduction}

Biomedical vision--language models (VLMs), such as BiomedCLIP~\citep{zhang2023biomedclip}, learn aligned visual and textual representations from large-scale biomedical image--text corpora and enable zero-shot image classification. The resulting representations provide a transferable foundation for downstream tasks, avoiding training from scratch for each target task. However, these general representations may not capture the subtle visual differences that distinguish classes in a target biomedical task~\citep{zhao2025clipmedical}. Collecting sufficient labeled examples for full supervision is costly because biomedical annotation often requires specialized clinical expertise~\citep{huang2023rethinking}. Few-shot adaptation addresses this constraint by adapting the pretrained model with a small labeled support set~\citep{shakeri2024fewshot}.

\begin{figure}[t]
\centering
\includegraphics[width=\columnwidth]{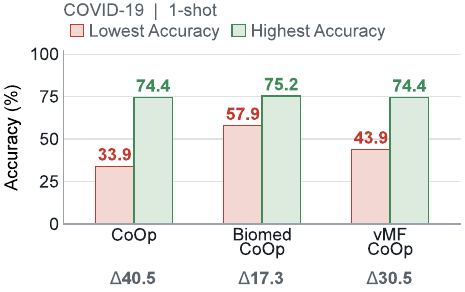}
\caption{Lowest and highest Accuracy across 20 support draws for three prompt
learning methods on COVID-19.}
\label{fig:support-selection-case}
\end{figure}

In few-shot adaptation, the labeled support examples provide the only supervision from the target task. Given their limited number, the sampled examples capture only part of the variation within the target classes. The adapted predictor may therefore reflect both task-relevant information and characteristics specific to the selected examples~\citep{xu2022alleviating}. Figure~\ref{fig:support-selection-case} illustrates the practical consequence of this dependence. For the same COVID-19 task, the lowest and highest Accuracy among 20 support draws are 33.9\% and 74.4\% for CoOp, with BiomedCoOp and vMFCoOp spanning 17.3\% and 30.5\% on the same draws. Reliable few-shot adaptation therefore depends on extracting task-specific information without making predictions overly dependent on the composition of a particular support set.

Prompt learning is a widely used approach to few-shot VLM adaptation. It keeps the pretrained encoders frozen and learns continuous context vectors from labeled support examples~\citep{zhou2021coop,zhou2022cocoop}. Biomedical extensions incorporate domain knowledge, semantic priors, or additional constraints to improve the task specificity of the learned prompts~\citep{koleilat2024biomedcoop,shao2025vmfcoop}. Other approaches adapt the visual representation or classifier while retaining pretrained text knowledge through low-rank updates, residual learning, or prototype mixing~\citep{zanella2024cliplora,yu2022taskres,silvarodriguez2023clap,silvarodriguez2025sstext,goswami2026tamp}. Across these formulations, the adapted prediction is determined by the sampled support set. Since this small set captures only part of the visual variation within the target classes, the learned predictor can be shaped by characteristics specific to the selected examples.

The frozen prompt prediction is unaffected by support selection and reflects the semantic knowledge acquired during pretraining~\citep{radford2021clip,zhang2023biomedclip}. It therefore complements the task information learned from the support set. Building on this contrast, we propose Prompt-Anchored Residual Adaptation (PARA). PARA retains the frozen prompt prediction as an anchor and incorporates a visual prediction learned from the support set as a residual relative to it. The visual prediction is obtained from a visual encoder adapted through low-rank updates and class prototypes that combine text anchors with adapted support representations. The residual step is computed in a closed form from frozen support embeddings using anchor discrepancy and support agreement. Anchor discrepancy measures the difference between the support and text directions, while support agreement measures consistency among support examples. The same global bound is used across all datasets, without per-dataset tuning or a validation set.

The influence of the support set on adaptation outcomes also raises a separate evaluation issue. Biomedical prompt-learning benchmarks report the mean and standard deviation over three random runs with independently sampled support sets~\citep{koleilat2024biomedcoop,shao2025vmfcoop}. This evaluation does not separate support-selection variation from optimization randomness or characterize lower-tail performance. We therefore evaluate 20 support sets for each dataset and shot level and use the same draws for all methods. Our evaluation isolates support selection by aggregating repeated optimization runs within each draw before computing support-level summaries. For both Accuracy and Macro-F1, we report the mean and CVaR$_{20}$ across support draws, where CVaR$_{20}$ is the average performance over the worst 20\% of draws~\citep{rockafellar2000optimization}. This lower tail is practically relevant when adaptation relies on a single available support set because it characterizes performance across the worst 20\% of draws rather than only average performance.

Our main contributions are summarized as follows:
\begin{itemize}
\item We propose Prompt-Anchored Residual Adaptation, which retains the frozen prompt prediction as an anchor and incorporates a visual prediction learned from the support set through an anchor-relative residual. The residual step is determined by frozen support geometry.
\item We introduce an evaluation protocol for fair and reliable comparison under support selection. It evaluates all methods on repeated, shared support draws and reports both the mean and CVaR$_{20}$ to characterize overall and lower-tail performance.
\item Across 11 biomedical datasets and five shot levels, PARA achieves the highest mean and CVaR$_{20}$ results for both Accuracy and Macro-F1, together with the strongest base-to-novel generalization.
\end{itemize}

\section{Related Work}

\subsection{Biomedical Vision-Language Adaptation}

Vision-language pretraining, exemplified by CLIP~\citep{radford2021clip}, has
been extended to biomedical image analysis. PubMedCLIP adapts CLIP to
medical visual question answering~\citep{eslami2021pubmedclip}, MedCLIP
learns from unpaired medical images and text via semantic
matching~\citep{wang2022medclip}. BiomedCLIP scales this recipe to large-scale
biomedical figure-caption pairs and transfers across retrieval, zero-shot
classification, and medical visual question
answering~\citep{zhang2023biomedclip}. These models instantiate a zero-shot
classifier from class names, but the resulting
classifier inherits the prompt form and the pretrained semantic space, which
may not capture the visual distinctions required by a target classification
task.

Few-shot adaptation addresses this mismatch with limited supervision. Prompt
learning for natural-image CLIP replaces hand-crafted templates with learnable
context tokens, conditions them on image features, aligns their updates with
pretrained knowledge, or extends prompting to both
branches~\citep{zhou2021coop,zhou2022cocoop,zhu2023prograd,khattak2022maple}.
Biomedical methods build on this route: BiomedCoOp regularizes context learning
with LLM-generated prompt ensembles~\citep{koleilat2024biomedcoop}, vMFCoOp
estimates class representations on a unified hyperspherical
manifold~\citep{shao2025vmfcoop}, and BioDPP learns an image-conditioned prompt
policy~\citep{miao2026biodpp}. Other methods act on the visual representation
instead: CLIP-Adapter attaches lightweight feature
adapters~\citep{gao2021clipadapter}, Tip-Adapter builds a cache classifier from
the support examples~\citep{zhang2022tipadapter}, and visual prompt tuning and
low-rank updates introduce trainable components into the visual
pathway~\citep{jia2022vpt,zanella2024cliplora}.
Whichever component is adapted, the task-specific signal is read from the
support set, so the resulting prediction is conditioned on the particular
labeled examples drawn.

\begin{figure*}[t]
\centering
\includegraphics[width=\textwidth]{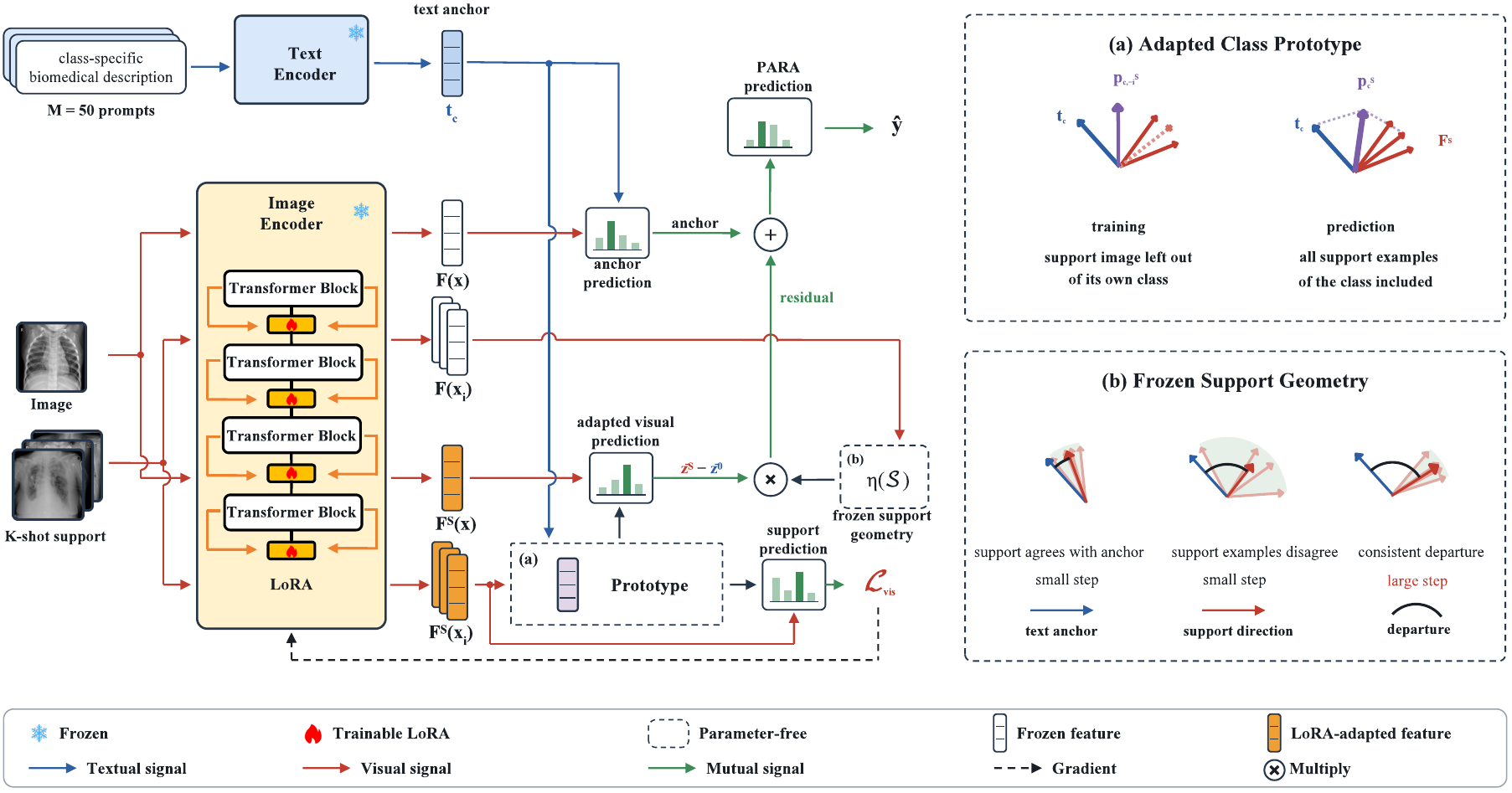}
\caption{Overview of PARA. During training, the support set is used to optimize the LoRA parameters. At inference, all support examples form the adapted class prototypes, and a query image is scored by both the frozen prompt branch and the adapted visual branch. Their difference is scaled by $\eta(\mathcal S)$ and added to the anchor prediction, where $\eta(\mathcal S)$ aggregates class-wise anchor discrepancy and support agreement computed from frozen support embeddings. Panel (a) details prototype construction during training and inference, while panel (b) illustrates how the two geometry quantities jointly determine the residual step.}
\label{fig:method-overview}
\end{figure*}

\subsection{Anchoring Adaptation to Frozen Text Knowledge}

A separate line of work preserves frozen text knowledge within the adapted predictor. TaskRes freezes the text classifier and adds a separately parameterized task residual, using a preset scale in its standard formulation~\citep{yu2022taskres}. LP++ forms each classifier weight by adding a class-wise scaled text embedding to a learned visual prototype, with both terms optimized on the support set~\citep{huang2024lpplusplus}. CLAP initializes a linear probe from the zero-shot prototypes and constrains its departure from them, with class-wise multipliers derived from zero-shot confidence on the support examples through an augmented-Lagrangian formulation~\citep{silvarodriguez2023clap}. For medical VLMs, SS-Text+ constructs class prototypes analytically and adjusts the relative contributions of text and visual evidence according to the number of support examples per class~\citep{silvarodriguez2025sstext}. TAMP projects image prototypes into a text-aligned subspace, mixes them with text prototypes under a shrinkage-estimation view, and complements the resulting classifier with an image-space LDA branch~\citep{goswami2026tamp}. These methods incorporate frozen text knowledge into an adapted classifier or its training objective. PARA retains the frozen prompt prediction as a separate branch and determines an anchor-relative residual step from the frozen geometry of the current support set.

\subsection{Evaluation under Support-Set Sampling}

Few-shot adaptation is commonly summarized over a small number of repeated runs. Biomedical prompt-learning benchmarks report the mean and standard deviation over three sampled support sets~\citep{koleilat2024biomedcoop,shao2025vmfcoop}. This practice recognizes support sampling, but three draws provide only a limited view of the support-level distribution and are insufficient to characterize its lower tail. The reported standard deviation also does not separate variation due to support selection from method-internal optimization randomness. \citet{bouthillier2021accounting} show that benchmark conclusions depend on which sources of variation are randomized and that data sampling contributes substantially to the observed spread. In few-shot adaptation, resampling the support set replaces the labeled data from which the task-specific predictor is built. Our protocol therefore evaluates 20 shared support draws, aggregates repeated optimization within each draw, and reports both mean performance and CVaR$_{20}$ over the worst 20\% of draws.

\section{Method}
\label{sec:method}

Figure~\ref{fig:method-overview} summarizes Prompt-Anchored Residual
Adaptation. PARA separates learning task-specific visual evidence from
deciding how far the prediction should move toward it.

\begin{table*}[t]
\centering
\scriptsize
\setlength{\tabcolsep}{1.2pt}
\newcommand{\tabres}[2]{#1\,{\tiny$\pm$\,#2}}
\begin{tabularx}{\textwidth}{@{}l*{12}{>{\centering\arraybackslash}X}@{}}
\toprule
& \multicolumn{6}{c}{\textbf{Accuracy}}
& \multicolumn{6}{c}{\textbf{Macro-F1}} \\
\cmidrule(lr){2-7}\cmidrule(lr){8-13}
\textbf{Method}
& $K=1$ & $K=2$ & $K=4$ & $K=8$ & $K=16$ & \textbf{Avg.}
& $K=1$ & $K=2$ & $K=4$ & $K=8$ & $K=16$ & \textbf{Avg.} \\
\midrule
\rowcolor{gray!18}
\multicolumn{13}{c}{\textbf{Zero-shot Method}} \\
BiomedCLIP
& \multicolumn{6}{c}{54.35}
& \multicolumn{6}{c}{44.55} \\
\midrule
\rowcolor{gray!18}
\multicolumn{13}{c}{\textbf{CLIP-Based Adaptation Methods}} \\
Tip-Adapter
& \tabres{55.12}{0.60} & \tabres{55.99}{0.81} & \tabres{57.16}{1.10}
& \tabres{59.22}{1.45} & \tabres{61.91}{1.42} & \tabres{57.88}{1.08}
& \tabres{45.46}{0.53} & \tabres{46.24}{0.78} & \tabres{47.63}{0.89}
& \tabres{49.89}{1.25} & \tabres{52.91}{1.52} & \tabres{48.43}{1.00} \\
CLIP-LoRA
& \tabres{54.82}{1.18} & \tabres{54.82}{1.09} & \tabres{55.23}{0.91}
& \tabres{56.15}{1.15} & \tabres{59.71}{1.40} & \tabres{56.15}{1.15}
& \tabres{46.53}{0.92} & \tabres{47.07}{0.86} & \tabres{47.77}{0.76}
& \tabres{50.46}{0.88} & \tabres{55.33}{1.20} & \tabres{49.43}{0.92} \\
TaskRes
& \tabres{53.89}{7.36} & \tabres{58.53}{5.84} & \tabres{62.33}{4.03}
& \tabres{67.02}{3.09} & \tabres{70.75}{2.27} & \tabres{62.50}{4.52}
& \tabres{49.24}{6.02} & \tabres{54.48}{4.80} & \tabres{59.09}{3.39}
& \tabres{64.07}{2.54} & \tabres{67.93}{1.97} & \tabres{58.96}{3.75} \\
CLAP
& \tabres{47.74}{9.41} & \tabres{55.69}{6.66} & \tabres{61.20}{4.89}
& \tabres{66.72}{3.22} & \tabres{70.53}{2.40} & \tabres{60.38}{5.32}
& \tabres{43.46}{7.34} & \tabres{51.83}{5.61} & \tabres{58.04}{4.09}
& \tabres{63.82}{2.61} & \tabres{67.85}{2.12} & \tabres{57.00}{4.36} \\
LP++
& \tabres{53.99}{7.13} & \tabres{58.48}{6.03} & \tabres{62.43}{4.21}
& \tabres{66.84}{3.16} & \tabres{70.58}{2.32} & \tabres{62.46}{4.57}
& \tabres{49.32}{5.79} & \tabres{54.49}{4.96} & \tabres{59.21}{3.52}
& \tabres{63.92}{2.64} & \tabres{67.81}{2.02} & \tabres{58.95}{3.79} \\
SS-Text+
& \tabres{49.32}{8.68} & \tabres{56.41}{6.51} & \tabres{61.40}{5.07} & \tabres{65.26}{3.85}
& \tabres{68.10}{2.72} & \tabres{60.10}{5.37} & \tabres{45.04}{7.28} & \tabres{52.46}{5.36}
& \tabres{58.11}{3.96} & \tabres{62.23}{3.03} & \tabres{65.06}{2.23} & \tabres{56.58}{4.37} \\
TAMP
& \tabres{54.53}{0.72} & \tabres{56.47}{6.70} & \tabres{62.66}{5.03}
& \tabres{67.91}{3.32} & \tabres{72.11}{2.25} & \tabres{62.74}{3.60}
& \tabres{44.65}{0.68} & \tabres{52.46}{5.57} & \tabres{59.48}{4.04}
& \tabres{65.05}{2.64} & \tabres{69.41}{1.96} & \tabres{58.21}{2.98} \\
\midrule
\rowcolor{gray!18}
\multicolumn{13}{c}{\textbf{Prompt Learning Methods}} \\
CoOp
& \tabres{50.12}{6.95} & \tabres{54.75}{5.03} & \tabres{57.74}{4.13}
& \tabres{61.82}{2.91} & \tabres{66.67}{2.50} & \tabres{58.22}{4.30}
& \tabres{46.13}{5.60} & \tabres{50.99}{4.00} & \tabres{54.34}{3.45}
& \tabres{57.99}{2.38} & \tabres{62.96}{2.04} & \tabres{54.48}{3.49} \\
CoCoOp
& \tabres{47.36}{4.40} & \tabres{50.95}{3.88} & \tabres{54.99}{3.47}
& \tabres{59.92}{2.84} & \tabres{64.36}{2.50} & \tabres{55.52}{3.42}
& \tabres{42.52}{3.39} & \tabres{46.61}{2.95} & \tabres{50.70}{2.72}
& \tabres{55.88}{2.18} & \tabres{60.39}{1.82} & \tabres{51.22}{2.61} \\
ProGrad
& \tabres{50.23}{7.13} & \tabres{56.18}{5.03} & \tabres{59.77}{4.05}
& \tabres{65.97}{3.14} & \tabres{70.63}{2.37} & \tabres{60.56}{4.34}
& \tabres{45.66}{6.01} & \tabres{51.55}{4.34} & \tabres{55.90}{3.43}
& \tabres{62.00}{2.57} & \tabres{67.14}{1.92} & \tabres{56.45}{3.65} \\
BiomedCoOp
& \tabres{56.89}{5.12} & \tabres{60.93}{3.82} & \tabres{65.16}{3.32}
& \tabres{69.01}{2.18} & \tabres{71.86}{1.61} & \tabres{64.77}{3.21}
& \tabres{48.75}{4.71} & \tabres{53.41}{3.60} & \tabres{57.79}{3.08}
& \tabres{61.70}{2.14} & \tabres{64.85}{1.54} & \tabres{57.30}{3.01} \\
vMFCoOp
& \tabres{52.84}{6.35} & \tabres{57.13}{5.10} & \tabres{60.89}{4.49}
& \tabres{65.48}{3.30} & \tabres{68.64}{2.80} & \tabres{61.00}{4.40}
& \tabres{47.64}{5.28} & \tabres{52.29}{4.00} & \tabres{56.53}{3.70}
& \tabres{61.26}{2.59} & \tabres{64.48}{2.15} & \tabres{56.44}{3.55} \\
\midrule
\rowcolor{blue!10}
\textbf{PARA}
& \textbf{\tabres{58.70}{3.30}} & \textbf{\tabres{62.11}{3.49}}
& \textbf{\tabres{66.10}{3.78}} & \textbf{\tabres{71.80}{2.64}}
& \textbf{\tabres{76.03}{2.09}} & \textbf{\tabres{66.95}{3.06}}
& \textbf{\tabres{50.17}{3.01}} & \textbf{\tabres{54.85}{3.31}}
& \textbf{\tabres{60.89}{3.33}} & \textbf{\tabres{67.81}{2.63}}
& \textbf{\tabres{72.95}{1.95}} & \textbf{\tabres{61.34}{2.84}} \\
\bottomrule
\end{tabularx}
\caption{Comparison with state-of-the-art methods on 11 biomedical datasets (\%). Entries report repeated-support means $\pm$ average within-dataset support-draw standard deviations. BiomedCLIP is support-invariant; best means are bold.}
\label{tab:main-few-shot}
\end{table*}

\subsection{Prompt-Anchored Visual Adaptation}
\label{sec:visual-adaptation}

A pretrained biomedical vision--language model provides a frozen visual
encoder $F_v^0$ and a frozen text encoder
$F_t$~\citep{zhang2023biomedclip}. We use $F_v^0(\cdot)$ and
$F_t(\cdot)$ to denote their $\ell_2$-normalized embeddings in
$\mathbb{R}^d$. For a downstream task with class set
$\mathcal{C}=\{1,\ldots,C\}$, each class $c\in\mathcal C$ is associated with
a fixed prompt ensemble $\{p_m(c)\}_{m=1}^{M}$~\citep{koleilat2024biomedcoop}.
Its \emph{text anchor} is the
normalized mean of the prompt embeddings,
\begin{equation}
  \mathbf{t}_c
  = \frac{\sum_{m=1}^{M}F_t\bigl(p_m(c)\bigr)}
         {\bigl\|\sum_{m=1}^{M}F_t\bigl(p_m(c)\bigr)\bigr\|}.
  \label{eq:text-anchor}
\end{equation}
A labeled support set $\mathcal S=\{(x_i,y_i)\}_{i=1}^{N}$ is available for
adaptation. We write $\mathcal S_c=\{i:y_i=c\}$. We insert
trainable LoRA parameters $\theta$ into the visual
encoder~\citep{hu2021lora,zanella2024cliplora}, yielding the unit-normalized
adapted representation $F_v^S(\cdot)$. Only $\theta$ and a visual logit scale
$\tau_v$ are optimized on the support set.

The adapted class prototype combines the fixed text anchor with the adapted
representations of its support images:
\begin{equation}
  \mathbf p_c^S
  =
  \frac{
    \mathbf t_c+\sum_{i\in\mathcal S_c}F_v^S(x_i)
  }{
    \left\|\mathbf t_c+\sum_{i\in\mathcal S_c}F_v^S(x_i)\right\|
  }.
  \label{eq:adapted-prototype}
\end{equation}

During training, however, including $F_v^S(x_i)$ in its own positive prototype
would let the objective be reduced by matching each support image to itself
rather than to its class. We therefore classify each support image using a
leave-one-out version of the prototype. Specifically, when
classifying $(x_i,y_i)$, we exclude $F_v^S(x_i)$ from its positive-class
prototype, while the prototypes of all other classes remain unchanged.
Denoting these training prototypes by $\mathbf p_{c,-i}^S$, we optimize
\begin{equation}
  \mathcal L_{\mathrm{vis}}(\theta,\tau_v)
  =\frac{1}{N}\sum_{i=1}^{N}
  \operatorname{CE}\!\left(
    \left\{
      \tau_v\left\langle F_v^S(x_i),\mathbf p_{c,-i}^S\right\rangle
    \right\}_{c\in\mathcal C},y_i
  \right).
  \label{eq:visual-training}
\end{equation}
For a one-shot class, the positive training prototype reduces to the fixed
text anchor, while the support image continues to update the visual adapter
through the classification objective.

For prediction, we construct $\mathbf p_c^S$ from all support examples using
the adapted visual encoder. Given a query image $x$, its visual score for class
$c$ is
\begin{equation}
  z_c^S(x)
  =\tau_v\left\langle F_v^S(x),\mathbf p_c^S\right\rangle.
  \label{eq:visual-logit}
\end{equation}

\subsection{Frozen Support Geometry}
\label{sec:support-geometry}

Given a support set $\mathcal S$, PARA computes anchor discrepancy $\delta_c$
and support agreement $\phi_c$ in the frozen embedding space to parameterize
the overall residual strength. Let $\mathbf u_i=F_v^0(x_i)$ denote
the normalized frozen representation of support image $x_i$. For class $c$,
its aggregate vector and visual direction are
\begin{equation}
  \mathbf h_c=\sum_{i\in\mathcal S_c}\mathbf u_i,
  \qquad
  \mathbf d_c=\frac{\mathbf h_c}{\|\mathbf h_c\|}.
  \label{eq:frozen-support-direction}
\end{equation}
Anchor discrepancy $\delta_c$ measures the departure of this visual direction
from the text anchor:
\begin{equation}
  \delta_c
  =1-\langle\mathbf d_c,\mathbf t_c\rangle
  =\frac{1}{2}\|\mathbf d_c-\mathbf t_c\|^2,
  \qquad \delta_c\in[0,2].
  \label{eq:anchor-discrepancy}
\end{equation}
A larger $\delta_c$ indicates a greater angular discrepancy between the class
direction exhibited by the support examples and the frozen text anchor.
However, $\delta_c$ only measures the departure of the aggregate direction
from the text anchor and does not capture agreement among the support
examples. To complement this information, support agreement $\phi_c$ is
defined from the class resultant~\citep{mardia2000directional,banerjee2005vmf}:
\begin{equation}
  \phi_c=\frac{\|\mathbf h_c\|^2}{n_c},
  \qquad n_c=|\mathcal S_c|,
  \qquad 0\leq\phi_c\leq n_c.
  \label{eq:support-agreement}
\end{equation}
Because the support representations are unit normalized, aligned
representations reinforce one another in $\mathbf h_c$, whereas divergent
representations cancel. Thus, $\phi_c$ reflects both the number of support
examples and their directional agreement. It equals one in the one-shot case
and reaches $n_c$ when all support representations are aligned.

\begin{table}[t]
\centering
\small
\setlength{\tabcolsep}{2pt}
\begin{tabularx}{\columnwidth}{@{}l*{6}{>{\centering\arraybackslash}X}@{}}
\toprule
& \multicolumn{5}{c}{\textbf{Shots per class ($K$)}} & \\
\cmidrule(lr){2-6}
\textbf{Method}
& 1 & 2 & 4 & 8 & 16 & \textbf{Avg.} \\
\midrule
\rowcolor{gray!18}
\multicolumn{7}{c}{\textbf{Zero-shot Method}} \\
BiomedCLIP
& \multicolumn{6}{c}{54.35} \\
\midrule
\rowcolor{gray!18}
\multicolumn{7}{c}{\textbf{CLIP-Based Adaptation Methods}} \\
Tip-Adapter
& \textbf{54.27} & 54.86 & 55.59 & 57.13 & 59.71 & 56.31 \\
CLIP-LoRA
& 53.06 & 53.31 & 53.92 & 54.50 & 57.67 & 54.49 \\
TaskRes
& 43.27 & 49.64 & 56.38 & 62.23 & 67.27 & 55.76 \\
CLAP
& 34.31 & 45.84 & 53.74 & 61.79 & 66.81 & 52.50 \\
LP++
& 43.73 & 49.43 & 56.16 & 62.05 & 67.00 & 55.67 \\
SS-Text+
& 36.28 & 46.83 & 53.45 & 59.12 & 63.86 & 51.91 \\
TAMP
& 53.52 & 46.41 & 54.83 & 62.94 & 68.62 & 57.27 \\
\midrule
\rowcolor{gray!18}
\multicolumn{7}{c}{\textbf{Prompt Learning Methods}} \\
CoOp
& 40.12 & 47.52 & 51.30 & 57.53 & 62.98 & 51.89 \\
CoCoOp
& 41.20 & 45.56 & 49.80 & 55.60 & 60.45 & 50.52 \\
ProGrad
& 39.65 & 48.75 & 53.64 & 61.05 & 67.05 & 54.03 \\
BiomedCoOp
& 49.08 & 55.00 & 60.12 & 65.79 & 69.48 & 59.89 \\
vMFCoOp
& 42.91 & 49.72 & 54.28 & 60.56 & 64.51 & 54.40 \\
\midrule
\rowcolor{blue!10}
\textbf{PARA}
& 53.78 & \textbf{57.04} & \textbf{60.39}
& \textbf{67.75} & \textbf{73.07} & \textbf{62.41} \\
\bottomrule
\end{tabularx}
\caption{Performance over the worst 20\% of support draws. Accuracy CVaR$_{20}$ (\%) is averaged across 11 biomedical datasets. BiomedCLIP is invariant to support selection. The best adaptation result in each column is shown in bold.}
\label{tab:lower-tail}
\end{table}

\begin{table*}[t]
\centering
\scriptsize
\setlength{\tabcolsep}{1.2pt}
\newcommand{\bntabres}[2]{#1\,{\tiny$\pm$\,#2}}
\begin{tabularx}{\textwidth}{@{}l
*{3}{>{\hsize=1.25\hsize\centering\arraybackslash}X}
*{3}{>{\hsize=0.75\hsize\centering\arraybackslash}X}
*{3}{>{\hsize=1.25\hsize\centering\arraybackslash}X}
*{3}{>{\hsize=0.75\hsize\centering\arraybackslash}X}@{}}
\toprule
& \multicolumn{6}{c}{\textbf{Accuracy}} & \multicolumn{6}{c}{\textbf{Macro-F1}} \\
\cmidrule(lr){2-7}\cmidrule(lr){8-13}
& \multicolumn{3}{c}{Mean $\pm$ SD} & \multicolumn{3}{c}{CVaR$_{20}$}
& \multicolumn{3}{c}{Mean $\pm$ SD} & \multicolumn{3}{c}{CVaR$_{20}$} \\
\cmidrule(lr){2-4}\cmidrule(lr){5-7}\cmidrule(lr){8-10}\cmidrule(lr){11-13}
\textbf{Method}
& Base & Novel & HM & Base & Novel & HM
& Base & Novel & HM & Base & Novel & HM \\
\midrule
BiomedCLIP
& 55.87 & 79.91 & 65.76
& 55.87 & 79.91 & 65.76
& 49.09 & 68.77 & 57.29
& 49.09 & 68.77 & 57.29 \\
CLIP-LoRA
& \bntabres{78.00}{2.23} & \bntabres{72.56}{4.47} & \bntabres{75.18}{2.76}
& 74.62 & 66.40 & 70.27
& \bntabres{76.41}{2.19} & \bntabres{66.43}{3.53} & \bntabres{71.07}{2.47}
& 73.09 & 61.48 & 66.78 \\
CoOp
& \bntabres{72.54}{2.75} & \bntabres{68.59}{5.07} & \bntabres{70.51}{3.47}
& 68.38 & 62.10 & 65.09
& \bntabres{70.94}{2.58} & \bntabres{56.00}{4.40} & \bntabres{62.59}{3.31}
& 66.98 & 49.75 & 57.09 \\
CoCoOp
& \bntabres{71.20}{2.94} & \bntabres{68.37}{5.56} & \bntabres{69.76}{3.69}
& 66.82 & 60.52 & 63.51
& \bntabres{69.54}{2.72} & \bntabres{54.48}{5.03} & \bntabres{61.10}{3.63}
& 65.42 & 47.47 & 55.02 \\
ProGrad
& \bntabres{73.47}{3.10} & \bntabres{68.35}{4.90} & \bntabres{70.82}{3.74}
& 68.95 & 61.28 & 64.89
& \bntabres{71.66}{3.05} & \bntabres{57.03}{3.99} & \bntabres{63.51}{3.19}
& 67.21 & 51.34 & 58.21 \\
BiomedCoOp
& \bntabres{75.13}{2.51} & \bntabres{74.71}{3.68} & \bntabres{74.92}{2.48}
& 71.35 & 69.55 & 70.44
& \bntabres{72.19}{2.49} & \bntabres{62.17}{3.35} & \bntabres{66.81}{2.39}
& 68.49 & 57.37 & 62.44 \\
vMFCoOp
& \bntabres{76.06}{2.79} & \bntabres{71.03}{5.56} & \bntabres{73.46}{3.58}
& 71.96 & 62.82 & 67.08
& \bntabres{73.80}{2.45} & \bntabres{61.85}{4.67} & \bntabres{67.30}{3.37}
& 70.07 & 54.90 & 61.56 \\
\midrule
\rowcolor{blue!10}
\textbf{PARA}
& \textbf{\bntabres{81.38}{2.59}} & \textbf{\bntabres{80.32}{0.26}} & \textbf{\bntabres{80.85}{1.36}}
& \textbf{77.47} & \textbf{79.96} & \textbf{78.70}
& \textbf{\bntabres{79.71}{2.58}} & \textbf{\bntabres{69.86}{0.30}} & \textbf{\bntabres{74.46}{1.17}}
& \textbf{75.83} & \textbf{69.46} & \textbf{72.51} \\
\bottomrule
\end{tabularx}
\caption{Base-to-novel generalization over 20 support selections on 10 datasets (\%). Adaptation uses 16-shot base-class support. SD is averaged across datasets. For both mean and CVaR$_{20}$, HM is computed from the displayed Base and Novel values; HM SD is computed from the corresponding draw-wise harmonic means.}
\label{tab:base-to-novel}
\end{table*}

\subsection{Anchor-Relative Residual Prediction}
\label{sec:residual}

The frozen prompt branch scores a query image $x$ as
\begin{equation}
  z_c^0(x)
  =\tau_0\bigl\langle F_v^0(x),\mathbf t_c\bigr\rangle,
  \label{eq:anchor-logit}
\end{equation}
where $\tau_0$ is the frozen pretrained logit scale. For a fixed query,
$z_c^0(x)$ is independent of both the sampled support set and adapter
optimization.

The text anchor and visual branches have independently learned logit scales. We
therefore standardize each score vector across the candidate classes, denoting
the resulting scores by $\bar z_c^0(x)$ and $\bar z_c^S(x)$. PARA expresses
the visual prediction as a residual relative to the frozen prompt prediction:
\begin{equation}
  \widehat z_c(x)
  =\bar z_c^0(x)
  +\eta(\mathcal S)\bigl(\bar z_c^S(x)-\bar z_c^0(x)\bigr).
  \label{eq:residual}
\end{equation}
The residual step is determined directly from the frozen support geometry:
\begin{equation}
  \eta(\mathcal S)
  =\frac{\lambda}{C}
  \sum_{c=1}^{C}
  \frac{\delta_c\phi_c}{1+\delta_c\phi_c}.
  \label{eq:residual-step}
\end{equation}
We use $\delta_c\phi_c$ as a soft conjunction: a class contributes strongly only when its support direction both departs from the text anchor and is consistently supported by its examples. We apply the monotone map $g(a)=a/(1+a)$ to bound each class contribution while preserving its ordering. Averaging over classes normalizes the step across different numbers of classes, and $\lambda$ sets its global upper bound. Consequently, $0\leq\eta(\mathcal S)<\lambda$. This step depends only on frozen support representations. The final prediction is $\widehat y(x)=\arg\max_{c\in\mathcal C}\widehat z_c(x)$.

\subsection{Base-to-Novel Residual Transport}
\label{sec:base-to-novel}

Let $\mathcal C=\mathcal B\cup\mathcal N$ comprise disjoint base and novel
class sets, with labeled support available only for
$\mathcal B$~\citep{zhou2022cocoop,khattak2022maple}. Base classes use the
PARA prediction defined in Eq.~\eqref{eq:residual}.

Novel classes lack support from which to construct class-specific visual
directions. We therefore extend the anchor-relative formulation to the frozen
embedding space by transferring the visual--text displacement revealed by the
base support set. For each base class $b$, this displacement is
\begin{equation}
  \boldsymbol\Delta_b=\mathbf d_b-\mathbf t_b,
  \qquad b\in\mathcal B.
  \label{eq:base-displacement}
\end{equation}
For novel class $n\in\mathcal N$, semantic similarity between frozen text
anchors determines how the base-class displacements are combined:
\begin{equation}
  w_{nb}
  =
  \frac{
    \exp\!\left(\beta\langle\mathbf t_n,\mathbf t_b\rangle\right)
  }{
    \sum_{j\in\mathcal B}
    \exp\!\left(\beta\langle\mathbf t_n,\mathbf t_j\rangle\right)
  },
  \label{eq:b2n-weights}
\end{equation}
Semantically closer base classes thus receive larger transfer weights. Applying
the weighted displacement to the original novel text anchor gives
\begin{equation}
  \mathbf t_n'
  =
  \frac{
    \mathbf t_n+
    \gamma\sum_{b\in\mathcal B}w_{nb}\boldsymbol\Delta_b
  }{
    \left\|
      \mathbf t_n+
      \gamma\sum_{b\in\mathcal B}w_{nb}\boldsymbol\Delta_b
    \right\|}.
  \label{eq:novel-anchor-transfer}
\end{equation}
Here $\beta$ controls semantic selectivity and $\gamma$ controls the transfer
magnitude. The adjusted anchor transfers the visual--text relationship observed
in the base support set to semantically related novel classes. The novel-class
score is then obtained from the frozen visual representation:
\begin{equation}
  z_n'(x)=\tau_0\left\langle F_v^0(x),\mathbf t_n'\right\rangle.
  \label{eq:novel-logit}
\end{equation}

\section{Experiments}

\subsection{Repeated-Support Evaluation Protocol}
\label{sec:repeated-support-evaluation}

Following the biomedical few-shot benchmark established by BiomedCoOp~\citep{koleilat2024biomedcoop}, we use the same 11 datasets, BiomedCLIP backbone, and shot levels. We extend its image-level evaluation with a repeated-support protocol. We construct one fixed class-stratified image-level split for each dataset, reserving 30\% of the images as the support pool and the remaining 70\% as the test set. No image appears in both partitions. At each shot level, we generate 20 class-balanced support draws while balancing the reuse of support-pool examples across draws. All methods share exactly the same split and support draws. Further details are provided in Appendix~\ref{sec:app-protocols}.

Prior biomedical prompt-learning benchmarks average three runs in which support selection and optimization randomness vary jointly~\citep{koleilat2024biomedcoop,shao2025vmfcoop}. In contrast, each stochastic method is run with three independent optimization seeds within each of our 20 support draws, yielding 60 runs per dataset--shot condition. The three scores are averaged to obtain one result for each draw, while deterministic methods produce one result directly. This nested design separates support-selection variation from optimization randomness and yields 20 support-level Accuracy and Macro-F1 values for each method, dataset, and shot level. We report their mean as typical performance and CVaR$_{20}$, defined as the mean of the four lowest values, to measure performance over the worst 20\% of support draws~\citep{rockafellar2000optimization}.

\subsection{Experimental Setup}

\paragraph{Datasets and backbone.}
We evaluate on 11 biomedical image classification datasets spanning MRI, CT,
X-ray, ultrasound, endoscopy, fundus photography, and
histopathology~\citep{koleilat2024biomedcoop,shao2025vmfcoop}. All compared
methods use BiomedCLIP ViT-B/16~\citep{zhang2023biomedclip} as the pretrained
backbone. Complete dataset and preprocessing
details are provided in Appendices~\ref{sec:app-protocols} and~\ref{sec:app-implementation}.

\paragraph{Few-shot classification.} Following established biomedical prompt-learning benchmarks~\citep{koleilat2024biomedcoop,shao2025vmfcoop}, we evaluate Accuracy and Macro-F1 at $K\in\{1,2,4,8,16\}$. Statistical significance is assessed across datasets using two-sided Wilcoxon signed-rank tests~\citep{wilcoxon1945individual}, with Holm correction over baseline comparisons separately for mean and CVaR$_{20}$ within each metric~\citep{holm1979simple}.

\paragraph{Base-to-novel generalization.} Following the standard base-to-novel evaluation~\citep{zhou2022cocoop,khattak2022maple}, we use the class splits from BiomedCoOp~\citep{koleilat2024biomedcoop}. Models adapt using 16-shot support from the base classes and are evaluated separately on the base and novel classes. We report Base, Novel, and harmonic-mean (HM) results for Accuracy and Macro-F1. BUSI is excluded because its novel split contains only one class.

\paragraph{Compared methods.}
We compare PARA with zero-shot BiomedCLIP using a frozen prompt
ensemble~\citep{zhang2023biomedclip,koleilat2024biomedcoop} and representative
few-shot adaptation methods. Visual and classifier adaptation baselines include
Tip-Adapter~\citep{zhang2022tipadapter}, CLIP-LoRA~\citep{zanella2024cliplora},
TaskRes~\citep{yu2022taskres}, CLAP~\citep{silvarodriguez2023clap},
LP++~\citep{huang2024lpplusplus}, SS-Text+~\citep{silvarodriguez2025sstext},
and TAMP~\citep{goswami2026tamp}. Prompt-learning baselines include
CoOp~\citep{zhou2021coop},
CoCoOp~\citep{zhou2022cocoop}, ProGrad~\citep{zhu2023prograd},
BiomedCoOp~\citep{koleilat2024biomedcoop}, and
vMFCoOp~\citep{shao2025vmfcoop}.
On novel classes, Tip-Adapter and TAMP reduce to the frozen prompt prediction,
since their support statistics cover only base classes. We therefore exclude
them from the base-to-novel comparison.

\paragraph{Implementation details.}
To keep text-side supervision consistent, PARA, frozen BiomedCLIP, and all baselines that use fixed text prototypes (Tip-Adapter, CLIP-LoRA, TaskRes, CLAP, LP++, SS-Text+, and TAMP) construct those prototypes from the same 50 GPT-4-generated descriptions per class released with BiomedCoOp~\citep{koleilat2024biomedcoop}. Prompt-learning baselines retain their original prompt formulations. Published method-specific configurations are retained for the baselines, with final checkpoints evaluated without validation or test-driven selection. PARA inserts rank-8 LoRA modules
with $\alpha=16$ into the query, key, value, and output projections of the
visual encoder's attention layers. During training, support images are
augmented using random resized crops and horizontal flips. We optimize the
visual branch for a fixed 80 epochs with
AdamW~\citep{loshchilov2019decoupled}, using learning rates of $10^{-4}$ and
$10^{-3}$ for the LoRA parameters and visual logit scale, respectively, and a
weight decay of $10^{-4}$. We evaluate the final checkpoint without
validation-based model selection. We fix $\lambda=0.9$ across all experiments. For base-to-novel residual transport, we use fixed global settings $(\beta,\gamma)=(20,0.2)$, with no dataset-specific or validation-based tuning. Additional implementation details are provided in Appendix~\ref{sec:app-implementation}.

\begin{figure}[t]
\centering
\includegraphics[width=\columnwidth]{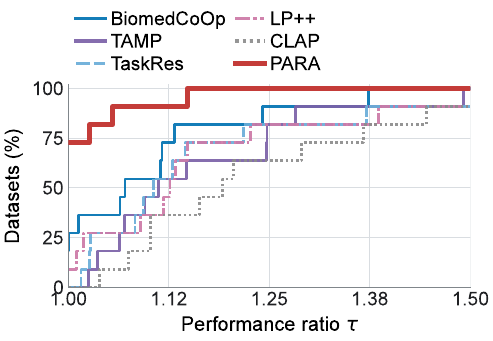}
\caption{Accuracy performance profiles for PARA and the five baselines used in paired significance tests across 11 datasets. A curve gives the fraction of datasets on which a method is within a factor $\tau$ of the lowest error among the methods shown~\citep{dolan2002benchmarking}.}
\label{fig:performance-profile}
\end{figure}

\newpage
\subsection{Main Results}

\subsubsection{Few-Shot Classification}

As shown in Table~\ref{tab:main-few-shot}, PARA achieves the highest Accuracy and Macro-F1 at all shot levels. Averaged across shot levels, PARA reaches 66.95\% Accuracy and 61.34\% Macro-F1, improving over the strongest competing results by 2.18\% and 2.38\%, respectively.

Figure~\ref{fig:performance-profile} uses performance profiles to summarize consistency. At $\tau=1$, each curve gives the fraction of datasets on which a method attains the lowest classification error; for $\tau>1$, it gives the fraction whose error is within a factor $\tau$ of the lowest error. PARA attains the lowest error on 8 of 11 datasets and is within a factor of 1.15 on all 11 datasets, showing that its aggregate gain is not driven by a small subset.

Table~\ref{tab:lower-tail} shows that PARA achieves the highest average Accuracy CVaR$_{20}$ of 62.41\%, outperforming the strongest competing result by 2.52\%.

\subsubsection{Base-to-Novel Generalization}

\begin{figure}[t]
\centering
\includegraphics[width=\columnwidth]{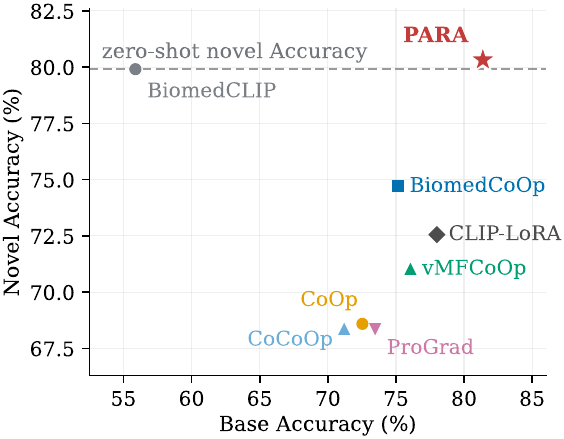}
\caption{The base--novel trade-off. Base and novel Accuracy (\%)
averaged over 10 datasets and 20 support selections. }
\label{fig:b2n-tradeoff}
\end{figure}

\begin{table}[t]
\centering
\small
\setlength{\tabcolsep}{1mm}
\begin{tabularx}{\columnwidth}{@{}>{\raggedright\arraybackslash}Xrrr@{}}
\toprule
\textbf{Variant} & \textbf{Mean Acc.} & \textbf{CVaR$_{20}$} & \textbf{SD} \\
\midrule
w/o Prototype Anchoring & 66.66 & 61.70 & 3.27 \\
w/o Prompt-Anchored Residual & 64.98 & 57.75 & 4.84 \\
w/o Support Geometry & 64.78 & 60.17 & 3.17 \\
\rowcolor{blue!10}
PARA (Full) & \textbf{66.95} & \textbf{62.41} & \textbf{3.06} \\
\bottomrule
\end{tabularx}
\caption{Component ablation. SD denotes the standard deviation across support draws.}
\label{tab:ablation}
\end{table}

Table~\ref{tab:base-to-novel} summarizes the complete base-to-novel results. PARA achieves the highest HM for both Accuracy (80.85\%) and Macro-F1 (74.46\%), improving over the best competing results of 75.18\% and 71.07\%, respectively. The same pattern holds for CVaR$_{20}$, where PARA achieves the highest HM for both metrics.

Figure~\ref{fig:b2n-tradeoff} further illustrates the base--novel trade-off in mean Accuracy. PARA achieves the highest Base Accuracy while preserving the novel-class performance of frozen BiomedCLIP. Its novel-class prediction uses the frozen-space anchor transport in Eq.~\eqref{eq:novel-anchor-transfer}. Together, these results indicate that the gains on the base classes do not come at the expense of novel-class performance. Complete per-dataset results are provided in Appendix~\ref{sec:app-base-to-novel}.

\begin{figure}[t]
\centering
\includegraphics[width=\columnwidth]{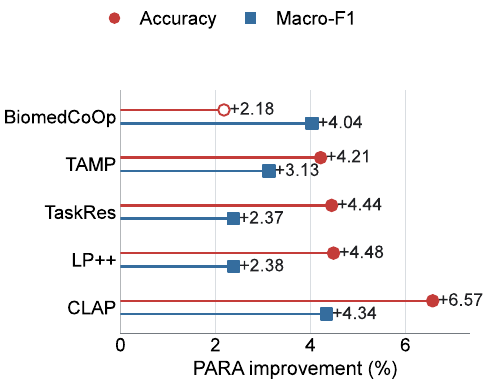}
\caption{Paired Wilcoxon tests. Filled markers indicate Holm-adjusted $p<0.05$.}
\label{fig:fewshot-significance}
\end{figure}

Figure~\ref{fig:fewshot-significance} shows that paired Wilcoxon tests yield significant gains in both metrics over all tested baselines except BiomedCoOp, for which only Macro-F1 is significant; complete per-dataset results and mean/CVaR$_{20}$ statistics are provided in Appendix~\ref{sec:app-few-shot}.

\subsubsection{Ablation Study}

Table~\ref{tab:ablation} evaluates PARA by removing each design in turn.
Without prototype anchoring, visual prototypes are formed only from support
features. Without the prompt-anchored residual, the adapted visual prediction
is used directly; without support geometry, the residual step is fixed at
$0.5$.

Removing prototype anchoring modestly reduces mean Accuracy and Accuracy
CVaR$_{20}$ from 66.95\% and 62.41\% to 66.66\% and 61.70\%. Removing the
prompt-anchored residual causes the largest lower-tail drop, reducing Accuracy
CVaR$_{20}$ to 57.75\% and increasing SD from 3.06 to 4.84. Using a fixed step
instead reduces mean Accuracy and Accuracy CVaR$_{20}$ to 64.78\% and 60.17\%.

\section{Conclusion}

We introduced Prompt-Anchored Residual Adaptation (PARA), which retains the frozen prompt prediction as an anchor and incorporates the adapted prediction as an anchor-relative residual. Its step is computed from anchor discrepancy and support agreement in frozen space. For fair and reliable comparison, our repeated-support protocol separates support-selection variation from optimization randomness and reports average and worst-20\% performance. Across 11 biomedical datasets, PARA achieves the strongest results on both measures while maintaining base-to-novel generalization. This perspective demonstrates that anchoring adaptation to frozen semantics improves both accuracy and stability.

\bibliography{main}

\clearpage
\onecolumn
\raggedbottom
\appendix
\setcounter{secnumdepth}{2}
\renewcommand{\thesection}{\Alph{section}}
\renewcommand{\thesubsection}{\thesection.\arabic{subsection}}
\setcounter{table}{0}
\setcounter{figure}{0}
\setcounter{equation}{0}
\renewcommand{\thetable}{A\arabic{table}}
\renewcommand{\thefigure}{A\arabic{figure}}
\renewcommand{\theequation}{A\arabic{equation}}
\renewcommand{\theproposition}{A\arabic{proposition}}
\section{Datasets and Evaluation Protocols}
\label{sec:app-protocols}

\subsection{Dataset Overview}

We evaluate on the 11 biomedical classification datasets used by the
BiomedCoOp benchmark~\citep{koleilat2024biomedcoop}: BTMRI~\citep{nickparvar2021btmri},
BUSI~\citep{aldhabyani2020busi}, CHMNIST~\citep{kather2016chmnist},
COVID-19~\citep{tahir2021covidquex}, CTKidney~\citep{islam2022ctkidney},
DermaMNIST~\citep{codella2019isic2018,tschandl2018ham10000},
KneeXray~\citep{chen2018kneeoa}, Kvasir~\citep{pogorelov2017kvasir},
LungColon~\citep{borkowski2019lc25000}, OCTMNIST~\citep{kermany2018oct},
and RETINA~\citep{kohler2013retina,porwal2018idrid}. Table~\ref{tab:app-datasets}
reports the modalities, numbers of classes, and image counts.

For every dataset and class, a seeded split stratified by class assigns 30\% of
the images to a support pool and the remaining 70\% to a fixed test set. The
two partitions are disjoint, and every method receives the same partition and
sample identities.

Images used by the frozen branch and at test time follow the deterministic
preprocessing supplied with BiomedCLIP ViT-B/16: conversion to RGB, resizing
and cropping to $224\times224$, and normalization with the pretrained image
statistics.

\subsection{Repeated Support Few-Shot Protocol}

For each dataset and shot level $K\in\{1,2,4,8,16\}$, we use 20 shared support
draws balanced across classes. Each draw contains exactly $K$ images per class,
and the reuse of examples from the support pool is balanced across draws.

Stochastic methods are evaluated with three optimizer seeds within every
support draw. Let $m_{d,s}$ denote a metric for support draw $d$ and optimizer
seed $s$. We first remove optimization randomness by computing
\begin{equation}
  \bar m_d=\frac{1}{3}\sum_{s=1}^{3}m_{d,s}.
\end{equation}
The reported cell mean and population standard deviation are then
\begin{equation}
  \mu=\frac{1}{20}\sum_{d=1}^{20}\bar m_d,
  \qquad
  \sigma=\sqrt{\frac{1}{20}\sum_{d=1}^{20}(\bar m_d-\mu)^2}.
\end{equation}
Deterministic methods contribute one value directly for each draw. To measure
performance in the lower tail, we sort the 20 draw values and define
\begin{equation}
  \operatorname{CVaR}_{20}=\frac{1}{4}\sum_{j=1}^{4}\bar m_{(j)},
\end{equation}
where $\bar m_{(1)}\leq\cdots\leq\bar m_{(20)}$. Thus, support draws rather
than optimizer runs are the statistical units. Accuracy and Macro-F1 are
computed on the common fixed test set; Macro-F1 is the mean of the
F1 scores for individual classes.

\subsection{Base-to-Novel Protocol}

Following BiomedCoOp~\citep{koleilat2024biomedcoop}, we adapt each method using
$K=16$ support from the base classes and evaluate it separately on the base
and novel classes. We report Base, Novel, and harmonic mean (HM) results for
Accuracy and Macro-F1. BUSI is excluded because its novel split contains only
one class, leaving ten evaluated datasets.

\begin{table*}[t]
\centering
\small
\setlength{\tabcolsep}{4pt}
\renewcommand{\arraystretch}{1.05}
\begin{tabular*}{\textwidth}{@{\extracolsep{\fill}}llrrrr@{}}
\toprule
Dataset & Modality & Classes & Total & Support pool & Test \\
\midrule
BTMRI & Magnetic Resonance Imaging & 4 & 5,712 & 1,714 & 3,998 \\
BUSI & Ultrasound & 3 & 780 & 234 & 546 \\
CHMNIST & Histopathology & 8 & 5,000 & 1,504 & 3,496 \\
COVID-19 & X-Ray & 4 & 21,165 & 6,351 & 14,814 \\
CTKidney & Computerized Tomography & 4 & 12,446 & 3,734 & 8,712 \\
DermaMNIST & Dermoscopy & 7 & 10,014 & 3,006 & 7,008 \\
KneeXray & X-Ray & 5 & 8,260 & 2,479 & 5,781 \\
Kvasir & Endoscopy & 8 & 4,000 & 1,200 & 2,800 \\
LungColon & Histopathology & 5 & 25,000 & 7,500 & 17,500 \\
OCTMNIST & Optical Coherence Tomography & 4 & 109,308 & 32,792 & 76,516 \\
RETINA & Fundus Photography & 4 & 4,217 & 1,264 & 2,953 \\
\bottomrule
\end{tabular*}
\caption{Dataset overview and the fixed split at the image level, stratified by class.}
\label{tab:app-datasets}
\end{table*}

\section{Implementation Details}
\label{sec:app-implementation}

We use the public BiomedCLIP~\citep{zhang2023biomedclip} ViT-B/16 checkpoint
and keep all pretrained image and text parameters frozen. The description bank
contains 50 descriptions generated by GPT-4 for each medical class. Following
BiomedCoOp, we use the $N=50$ descriptions released for every class.
BiomedCoOp generated them with the following instruction:

\begin{quote}\raggedright
Give $N$ textual descriptions of visual discriminative features\\
for distinct medical cases of \mbox{[CLASS]} found in \mbox{[MODALITY]}.
\end{quote}

We encode the 50 descriptions released with BiomedCoOp, normalize every text
embedding, average the 50 embeddings, and normalize the mean to obtain the
fixed text anchor $\mathbf t_c$.

\section{Complete Few-Shot Results}
\label{sec:app-few-shot}

\subsection{Results by Dataset}

\begin{figure*}[t]
  \centering
  \includegraphics[width=\textwidth]{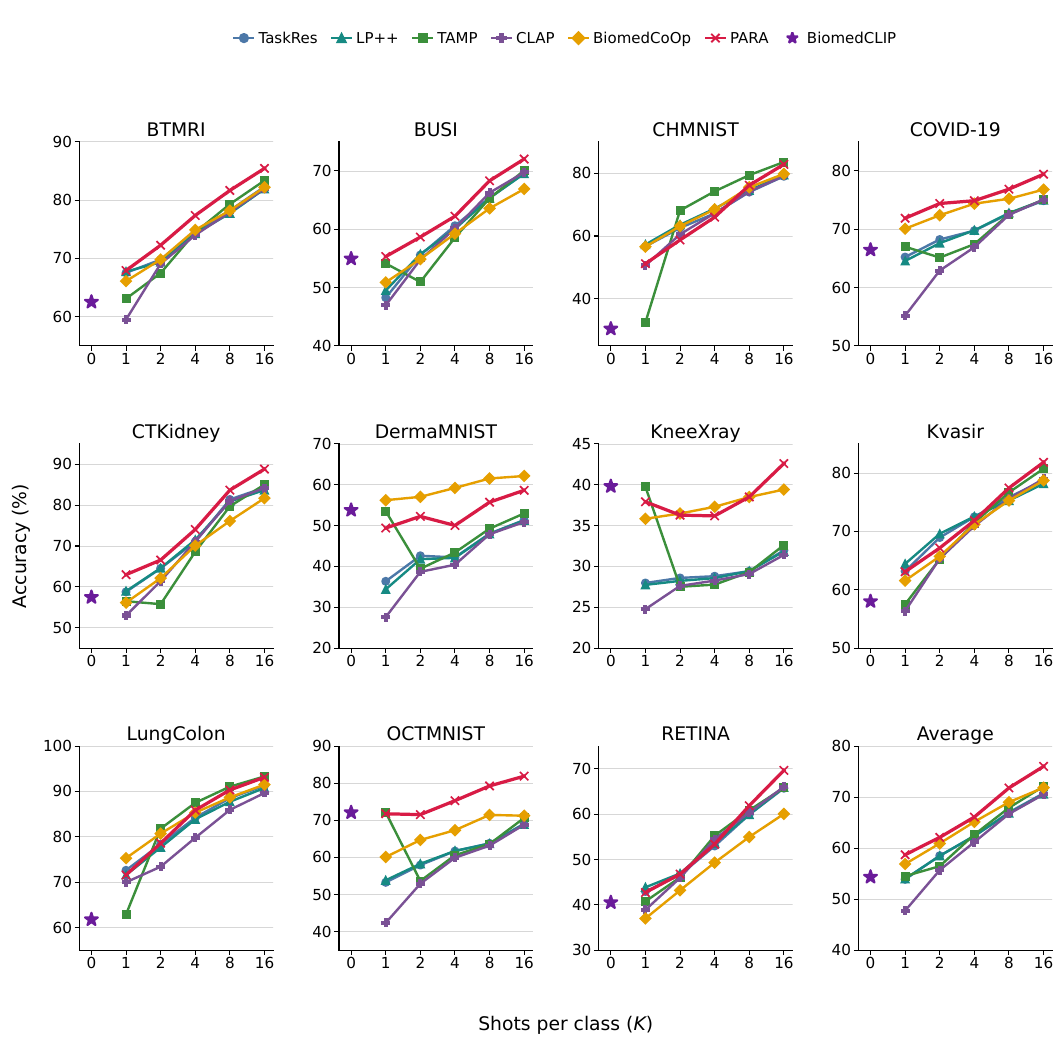}
  \caption{Few-shot Accuracy for each dataset. The Average panel is the mean
  across the 11 datasets at each shot level. Each adaptation point
  is the mean over 20 support draws after averaging optimizer seeds within
  each draw. BiomedCLIP denotes the zero-shot anchor at $K=0$, which is
  independent of the sampled support set.}
  \label{fig:app-fewshot-per-dataset-accuracy}
\end{figure*}

\begin{figure*}[t]
  \centering
  \includegraphics[width=\textwidth]{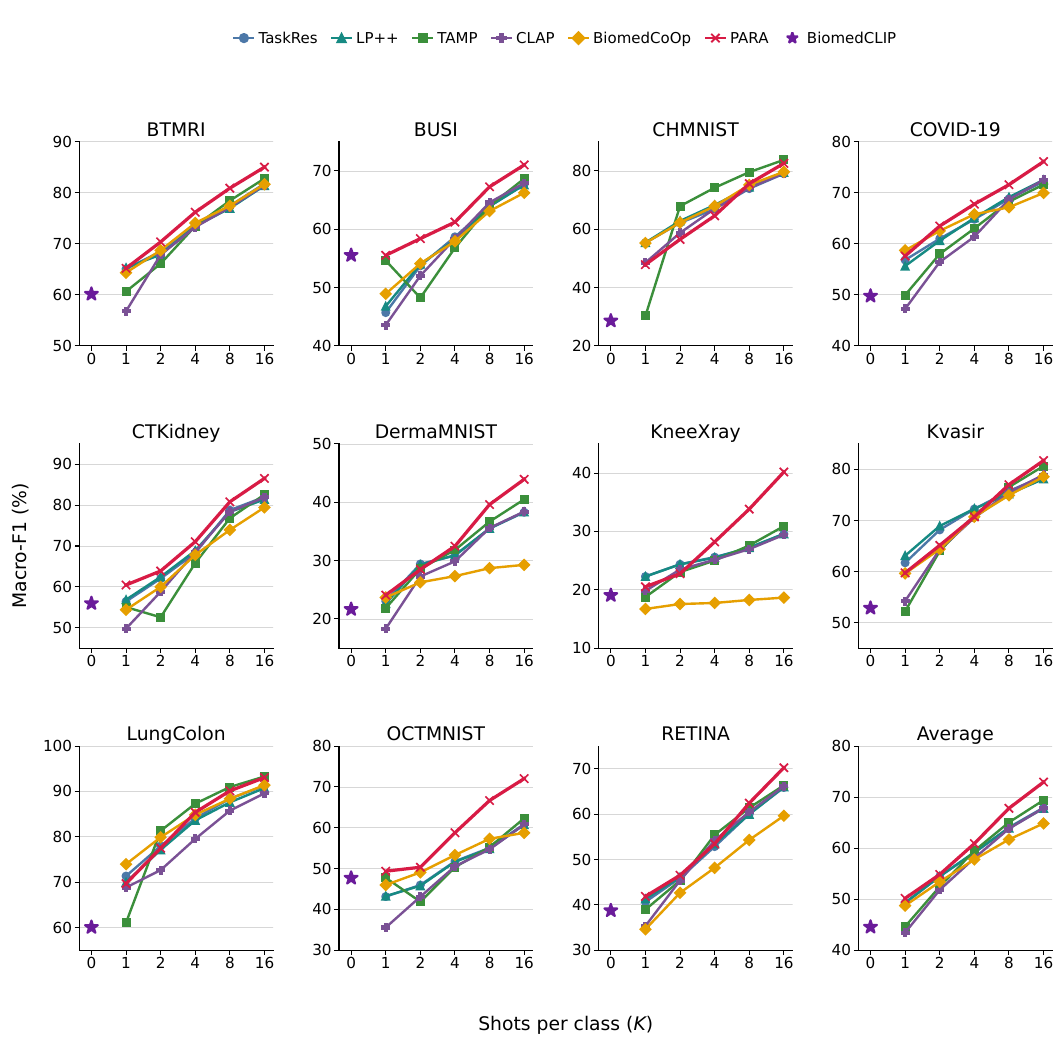}
  \caption{Few-shot Macro-F1 for each dataset under the same protocol as
  Figure~\ref{fig:app-fewshot-per-dataset-accuracy}. The Average panel is
  the mean across the 11 datasets at each shot level.}
  \label{fig:app-fewshot-per-dataset-macro-f1}
\end{figure*}

Figures~\ref{fig:app-fewshot-per-dataset-accuracy}
and~\ref{fig:app-fewshot-per-dataset-macro-f1} show mean Accuracy and
Macro-F1 at each shot level for every dataset and the average across datasets.
BiomedCLIP is plotted at $K=0$ because its frozen prediction is independent of
the sampled support set. The adaptation baselines are
Tip-Adapter~\citep{zhang2022tipadapter},
CLIP-LoRA~\citep{zanella2024cliplora},
TaskRes~\citep{yu2022taskres}, CLAP~\citep{silvarodriguez2023clap},
LP++~\citep{huang2024lpplusplus},
SS-Text+~\citep{silvarodriguez2025sstext}, TAMP~\citep{goswami2026tamp},
CoOp~\citep{zhou2021coop}, CoCoOp~\citep{zhou2022cocoop},
ProGrad~\citep{zhu2023prograd},
BiomedCoOp~\citep{koleilat2024biomedcoop}, and
vMFCoOp~\citep{shao2025vmfcoop}.
To keep the trajectories readable, the figures show PARA, TaskRes, LP++, TAMP,
CLAP, and BiomedCoOp. Table~\ref{tab:app-fewshot-dataset-summary} reports
per-dataset mean and CVaR$_{20}$ results for the complete comparison after
averaging each statistic over the five shot levels.

\subsection{Statistical Comparisons}

We assess paired differences across datasets. For each dataset at each shot
level, optimizer seeds are first averaged within each support draw, and the 20
draws are summarized. We then average the five shot levels within each dataset,
yielding 11 paired observations. Table~\ref{tab:app-wilcoxon} reports effect
summaries from exact two-sided Wilcoxon signed-rank
tests~\citep{wilcoxon1945individual}. The $p$-values are adjusted with Holm's
method~\citep{holm1979simple} over the 12 adaptation baselines separately
within each metric and summary.

\begin{table}[!b]
\centering
\footnotesize
\setlength{\tabcolsep}{2pt}
\renewcommand{\arraystretch}{1.03}
\begin{tabular*}{\textwidth}{@{\extracolsep{\fill}}lrrrrrr@{}}
\toprule
\multicolumn{7}{c}{\textbf{Accuracy}} \\
& \multicolumn{3}{c}{Mean} & \multicolumn{3}{c}{CVaR$_{20}$} \\
\cmidrule(lr){2-4}\cmidrule(lr){5-7}
Baseline & $\Delta$ & Wins & $p_{\rm H}$ & $\Delta$ & Wins & $p_{\rm H}$ \\
\midrule
Tip-Adapter & 9.07 & 9/11 & 0.0244 & 6.09 & 9/11 & 0.0645 \\
CLIP-LoRA & 10.80 & 11/11 & 0.0117 & 7.91 & 10/11 & 0.0195 \\
TaskRes & 4.44 & 10/11 & 0.0244 & 6.65 & 10/11 & 0.0137 \\
CLAP & 6.57 & 11/11 & 0.0117 & 9.91 & 11/11 & 0.0117 \\
LP++ & 4.48 & 10/11 & 0.0244 & 6.73 & 10/11 & 0.0176 \\
SS-Text+ & 6.85 & 9/11 & 0.0273 & 10.50 & 9/11 & 0.0205 \\
TAMP & 4.21 & 10/11 & 0.0205 & 5.14 & 10/11 & 0.0176 \\
CoOp & 8.73 & 11/11 & 0.0117 & 10.52 & 11/11 & 0.0117 \\
CoCoOp & 11.43 & 11/11 & 0.0117 & 11.88 & 11/11 & 0.0117 \\
ProGrad & 6.39 & 10/11 & 0.0205 & 8.38 & 11/11 & 0.0117 \\
BiomedCoOp & 2.18 & 8/11 & 0.1230 & 2.51 & 8/11 & 0.1748 \\
vMFCoOp & 5.95 & 11/11 & 0.0117 & 8.01 & 11/11 & 0.0117 \\
\bottomrule
\end{tabular*}
\caption{Paired comparisons between PARA and all 12 adaptation baselines across 11 datasets. $\Delta$ is the mean paired difference in percentage points, Wins counts positive differences, and $p_{\rm H}$ is adjusted with Holm's method over the 12 baselines separately for each summary. (a) Accuracy.}
\label{tab:app-wilcoxon}
\end{table}

\begin{table}[!t]
\ContinuedFloat
\centering
\footnotesize
\setlength{\tabcolsep}{2pt}
\renewcommand{\arraystretch}{1.03}
\begin{tabular*}{\textwidth}{@{\extracolsep{\fill}}lrrrrrr@{}}
\toprule
\multicolumn{7}{c}{\textbf{Macro-F1}} \\
& \multicolumn{3}{c}{Mean} & \multicolumn{3}{c}{CVaR$_{20}$} \\
\cmidrule(lr){2-4}\cmidrule(lr){5-7}
Baseline & $\Delta$ & Wins & $p_{\rm H}$ & $\Delta$ & Wins & $p_{\rm H}$ \\
\midrule
Tip-Adapter & 12.91 & 11/11 & 0.0117 & 10.17 & 11/11 & 0.0117 \\
CLIP-LoRA & 11.90 & 11/11 & 0.0117 & 9.09 & 11/11 & 0.0117 \\
TaskRes & 2.37 & 9/11 & 0.0391 & 3.76 & 9/11 & 0.0244 \\
CLAP & 4.34 & 10/11 & 0.0156 & 6.73 & 11/11 & 0.0117 \\
LP++ & 2.38 & 9/11 & 0.0410 & 3.87 & 9/11 & 0.0244 \\
SS-Text+ & 4.76 & 9/11 & 0.0420 & 7.19 & 9/11 & 0.0244 \\
TAMP & 3.13 & 10/11 & 0.0293 & 3.48 & 10/11 & 0.0244 \\
CoOp & 6.86 & 11/11 & 0.0117 & 7.92 & 11/11 & 0.0117 \\
CoCoOp & 10.12 & 11/11 & 0.0117 & 9.80 & 11/11 & 0.0117 \\
ProGrad & 4.89 & 9/11 & 0.0293 & 6.21 & 10/11 & 0.0137 \\
BiomedCoOp & 4.04 & 9/11 & 0.0410 & 4.42 & 10/11 & 0.0244 \\
vMFCoOp & 4.90 & 10/11 & 0.0205 & 6.01 & 10/11 & 0.0137 \\
\bottomrule
\end{tabular*}
\caption[]{Paired comparisons between PARA and all 12 adaptation baselines (continued). (b) Macro-F1.}
\end{table}

For mean Accuracy, PARA improves over 11 of the 12 baselines significantly;
the exception is BiomedCoOp ($p_{\rm H}=0.1230$). For Accuracy CVaR$_{20}$,
the gains are significant over 10 baselines, with Tip-Adapter
($p_{\rm H}=0.0645$) and BiomedCoOp ($p_{\rm H}=0.1748$) as the two
exceptions. Both mean and CVaR$_{20}$ Macro-F1 gains are significant over all
12 baselines. The remaining Accuracy losses are concentrated in a few
datasets, most notably DermaMNIST and CHMNIST, which are analyzed below.

\subsection{Failure Analysis}

\begingroup
The two main losses have different explanations. DermaMNIST is highly
imbalanced: melanocytic nevus accounts for 4,693/7,008 test images (66.97\%),
while the two smallest classes have only 80 and 99. BiomedCoOp has higher
mean/CVaR$_{20}$ Accuracy (59.23/54.90\% versus 53.21/43.24\%) but lower
Macro-F1 (27.05/25.30\% versus 33.71/30.03\%). This gap suggests majority class
bias in BiomedCoOp and more balanced PARA performance.

On CHMNIST, the frozen anchor has the lowest Accuracy among the 11 datasets
(30.43\%; Macro-F1 28.61\%), while the visual branch averages
75.28\%/74.84\% versus 66.96\%/65.45\% for final PARA in Accuracy/Macro-F1.
At one shot, the visual branch obtains 63.56\%/61.83\%, PARA using the step size
derived from support geometry obtains 51.18\%/47.87\%. As $K$
increases, the Accuracy gap between the visual branch and PARA falls
from 12.39 points at one shot to 2.89 points at 16 shots. On CHMNIST, the
visual branch is already strong, but the frozen text anchor is weak. Mixing
the two pulls PARA away from the stronger visual prediction and lowers its
final performance, especially at low shot counts.
\par\endgroup

\FloatBarrier
\subsection{Complete Result Tables}

\begin{table}[H]
\centering
\caption{Few-shot results for each dataset across all methods in the main comparison (\%). Each value is averaged over the five shot levels. Mean and CVaR$_{20}$ are shown in separate blocks. Bold marks the best result across all methods. (a) Frozen and CLIP-based adaptation methods.}
\label{tab:app-fewshot-dataset-summary}
\footnotesize
\setlength{\tabcolsep}{1.0pt}
\renewcommand{\arraystretch}{1.02}
\begin{tabular*}{\textwidth}{@{\extracolsep{\fill}}lrrrrrrrrr@{}}
\toprule
Dataset & BiomedCLIP & Tip-Adapter & CLIP-LoRA & TaskRes & CLAP & LP++ & SS-Text+ & TAMP & \textbf{PARA} \\
\midrule
\rowcolor{gray!12} \multicolumn{10}{l}{\textbf{Accuracy: Mean}} \\
\midrule
BTMRI & 62.53 & 65.79 & 61.95 & 74.49 & 72.50 & 74.19 & 71.36 & 73.54 & \textbf{76.93} \\
BUSI & 54.95 & 56.30 & 57.98 & 59.89 & 59.57 & 60.03 & 56.70 & 59.81 & \textbf{63.31} \\
CHMNIST & 30.43 & 41.86 & 44.95 & 67.88 & 66.34 & 68.69 & \textbf{71.63} & 67.55 & 66.96 \\
COVID-19 & 66.46 & 71.40 & 62.89 & 70.16 & 66.51 & 69.95 & 67.40 & 69.45 & \textbf{75.50} \\
CTKidney & 57.45 & 60.47 & 65.38 & 72.01 & 70.13 & 71.92 & 59.55 & 69.11 & \textbf{75.22} \\
DermaMNIST & 53.78 & 55.75 & 48.18 & 44.12 & 41.09 & 43.50 & 42.77 & 47.72 & 53.21 \\
KneeXray & 39.80 & \textbf{40.12} & 26.71 & 29.35 & 28.22 & 29.16 & 27.33 & 31.39 & 38.30 \\
Kvasir & 58.00 & 60.75 & 64.13 & 71.84 & 69.44 & 72.01 & 70.79 & 70.34 & \textbf{72.28} \\
LungColon & 61.77 & 69.01 & 72.09 & 83.00 & 79.76 & 82.32 & \textbf{85.72} & 83.31 & 83.90 \\
OCTMNIST & 72.11 & 73.09 & 70.16 & 61.17 & 57.44 & 61.30 & 56.93 & 64.13 & \textbf{75.95} \\
RETINA & 40.54 & 42.14 & 43.18 & 53.63 & 53.14 & 54.03 & 50.89 & 53.75 & \textbf{54.87} \\
\midrule
\rowcolor{gray!12} \multicolumn{10}{l}{\textbf{Accuracy: CVaR$_{20}$}} \\
\midrule
BTMRI & 62.53 & 64.57 & 60.63 & 68.94 & 65.26 & 69.00 & 62.95 & 68.50 & \textbf{72.71} \\
BUSI & 54.95 & 53.27 & 55.67 & 51.64 & 50.51 & 51.38 & 45.68 & 51.91 & \textbf{57.82} \\
CHMNIST & 30.43 & 40.99 & 43.92 & 64.20 & 61.33 & 65.12 & \textbf{67.08} & 64.21 & 63.73 \\
COVID-19 & 66.46 & 69.58 & 59.80 & 61.90 & 55.02 & 60.86 & 59.23 & 64.81 & \textbf{71.79} \\
CTKidney & 57.45 & 58.95 & 63.15 & 64.02 & 61.28 & 63.78 & 47.14 & 61.84 & \textbf{70.04} \\
DermaMNIST & 53.78 & 52.52 & 45.90 & 32.43 & 29.62 & 32.04 & 29.94 & 39.27 & 43.24 \\
KneeXray & \textbf{39.80} & 38.34 & 24.85 & 25.17 & 23.20 & 24.88 & 22.02 & 28.03 & 34.05 \\
Kvasir & 58.00 & 60.18 & 63.47 & 68.88 & 65.25 & 69.22 & 66.98 & 67.50 & \textbf{69.71} \\
LungColon & 61.77 & 68.26 & 70.60 & 78.12 & 73.96 & 78.31 & \textbf{81.49} & 80.54 & 81.00 \\
OCTMNIST & \textbf{72.11} & 71.62 & 69.23 & 50.53 & 46.22 & 50.27 & 44.41 & 54.98 & 71.72 \\
RETINA & 40.54 & 41.13 & 42.18 & 47.51 & 45.84 & 47.54 & 44.04 & 48.34 & \textbf{50.64} \\
\midrule
\rowcolor{gray!12} \multicolumn{10}{l}{\textbf{Macro-F1: Mean}} \\
\midrule
BTMRI & 60.15 & 63.31 & 59.61 & 73.32 & 71.31 & 73.01 & 70.11 & 72.30 & \textbf{75.54} \\
BUSI & 55.54 & 56.65 & 57.69 & 57.95 & 57.29 & 58.10 & 54.68 & 58.47 & \textbf{62.69} \\
CHMNIST & 28.61 & 38.71 & 43.13 & 67.48 & 65.55 & 68.09 & \textbf{71.40} & 67.16 & 65.45 \\
COVID-19 & 49.78 & 59.01 & 53.94 & 64.73 & 61.26 & 64.60 & 61.64 & 62.17 & \textbf{67.32} \\
CTKidney & 55.94 & 58.27 & 63.99 & 69.51 & 67.47 & 69.57 & 56.67 & 66.56 & \textbf{72.55} \\
DermaMNIST & 21.65 & 24.08 & 22.37 & 31.42 & 29.89 & 31.34 & 31.14 & 31.88 & \textbf{33.71} \\
KneeXray & 19.04 & 19.62 & 17.69 & 25.79 & 24.96 & 25.79 & 23.81 & 25.05 & \textbf{29.09} \\
Kvasir & 52.83 & 56.44 & 61.60 & 71.32 & 68.70 & \textbf{71.53} & 70.02 & 68.89 & 70.86 \\
LungColon & 60.06 & 67.51 & 70.61 & 82.61 & 79.28 & 81.81 & \textbf{85.26} & 82.79 & 83.15 \\
OCTMNIST & 47.68 & 48.95 & 49.83 & 51.37 & 48.95 & 51.26 & 47.10 & 51.51 & \textbf{59.43} \\
RETINA & 38.75 & 40.14 & 43.29 & 53.07 & 52.32 & 53.37 & 50.55 & 53.50 & \textbf{54.90} \\
\midrule
\rowcolor{gray!12} \multicolumn{10}{l}{\textbf{Macro-F1: CVaR$_{20}$}} \\
\midrule
BTMRI & 60.15 & 61.81 & 58.57 & 67.30 & 63.76 & 67.23 & 61.11 & 67.02 & \textbf{70.91} \\
BUSI & 55.54 & 53.86 & 55.49 & 50.32 & 48.67 & 49.53 & 44.35 & 51.37 & \textbf{57.94} \\
CHMNIST & 28.61 & 37.81 & 42.02 & 63.50 & 60.37 & 64.19 & \textbf{66.87} & 63.84 & 62.11 \\
COVID-19 & 49.78 & 57.02 & 52.04 & 58.32 & 52.47 & 57.69 & 55.89 & 58.43 & \textbf{62.17} \\
CTKidney & 55.94 & 56.56 & 61.89 & 61.91 & 58.80 & 62.08 & 45.44 & 59.63 & \textbf{66.95} \\
DermaMNIST & 21.65 & 23.50 & 21.76 & 26.61 & 25.09 & 26.39 & 26.32 & 28.52 & \textbf{30.03} \\
KneeXray & 19.04 & 18.90 & 16.19 & 22.71 & 21.51 & 22.64 & 20.28 & 22.58 & \textbf{26.35} \\
Kvasir & 52.83 & 55.64 & 60.72 & 68.31 & 64.10 & \textbf{68.60} & 65.67 & 65.70 & 67.76 \\
LungColon & 60.06 & 66.70 & 68.75 & 77.47 & 73.38 & 77.49 & \textbf{80.52} & 79.96 & 80.06 \\
OCTMNIST & 47.68 & 46.38 & 49.23 & 44.72 & 42.02 & 44.18 & 40.35 & 45.90 & \textbf{54.63} \\
RETINA & 38.75 & 38.97 & 42.36 & 46.42 & 44.82 & 46.45 & 43.12 & 47.81 & \textbf{50.09} \\
\bottomrule
\end{tabular*}
\end{table}

\begin{table}[H]
\ContinuedFloat
\centering
\caption[]{Few-shot results for each dataset (continued). (b) Prompt-learning methods; PARA is repeated for comparison.}
\footnotesize
\setlength{\tabcolsep}{1.0pt}
\renewcommand{\arraystretch}{1.02}
\begin{tabular*}{\textwidth}{@{\extracolsep{\fill}}lrrrrrr@{}}
\toprule
Dataset & CoOp & CoCoOp & ProGrad & BiomedCoOp & vMFCoOp & \textbf{PARA} \\
\midrule
\rowcolor{gray!12} \multicolumn{7}{l}{\textbf{Accuracy: Mean}} \\
\midrule
BTMRI & 72.93 & 68.06 & 72.92 & 74.24 & 74.66 & \textbf{76.93} \\
BUSI & 58.16 & 56.36 & 59.51 & 59.10 & 59.10 & \textbf{63.31} \\
CHMNIST & 64.30 & 58.07 & 66.77 & 68.69 & 66.53 & 66.96 \\
COVID-19 & 69.96 & 63.84 & 65.20 & 73.78 & 70.97 & \textbf{75.50} \\
CTKidney & 68.03 & 64.32 & 67.76 & 69.24 & 68.80 & \textbf{75.22} \\
DermaMNIST & 33.10 & 32.01 & 40.49 & \textbf{59.23} & 36.96 & 53.21 \\
KneeXray & 26.34 & 26.91 & 30.10 & 37.50 & 30.56 & 38.30 \\
Kvasir & 68.05 & 67.51 & 70.00 & 70.50 & 71.43 & \textbf{72.28} \\
LungColon & 81.03 & 77.06 & 84.19 & 84.31 & 83.47 & 83.90 \\
OCTMNIST & 56.66 & 58.29 & 63.15 & 66.97 & 61.58 & \textbf{75.95} \\
RETINA & 41.85 & 38.24 & 46.05 & 48.91 & 46.89 & \textbf{54.87} \\
\midrule
\rowcolor{gray!12} \multicolumn{7}{l}{\textbf{Accuracy: CVaR$_{20}$}} \\
\midrule
BTMRI & 67.23 & 64.16 & 65.56 & 68.80 & 69.98 & \textbf{72.71} \\
BUSI & 48.31 & 50.69 & 50.14 & 50.71 & 49.98 & \textbf{57.82} \\
CHMNIST & 60.31 & 54.85 & 63.33 & 64.72 & 62.98 & 63.73 \\
COVID-19 & 61.14 & 56.79 & 54.80 & 70.32 & 64.58 & \textbf{71.79} \\
CTKidney & 60.51 & 57.79 & 61.00 & 62.42 & 60.49 & \textbf{70.04} \\
DermaMNIST & 23.94 & 22.81 & 30.89 & \textbf{54.90} & 25.26 & 43.24 \\
KneeXray & 20.30 & 21.80 & 24.74 & 35.77 & 22.80 & 34.05 \\
Kvasir & 65.34 & 65.59 & 67.10 & 67.37 & 68.86 & \textbf{69.71} \\
LungColon & 77.20 & 74.23 & 80.28 & 80.34 & 79.72 & 81.00 \\
OCTMNIST & 50.62 & 52.55 & 55.76 & 61.10 & 52.59 & 71.72 \\
RETINA & 35.90 & 34.47 & 40.70 & 42.40 & 41.11 & \textbf{50.64} \\
\midrule
\rowcolor{gray!12} \multicolumn{7}{l}{\textbf{Macro-F1: Mean}} \\
\midrule
BTMRI & 71.92 & 65.73 & 71.88 & 73.25 & 73.47 & \textbf{75.54} \\
BUSI & 56.13 & 54.22 & 58.06 & 58.07 & 57.38 & \textbf{62.69} \\
CHMNIST & 63.67 & 56.85 & 65.86 & 68.07 & 65.51 & 65.45 \\
COVID-19 & 61.79 & 56.01 & 60.82 & 64.83 & 63.82 & \textbf{67.32} \\
CTKidney & 65.31 & 62.20 & 66.04 & 67.12 & 66.75 & \textbf{72.55} \\
DermaMNIST & 25.84 & 22.54 & 29.32 & 27.05 & 26.96 & \textbf{33.71} \\
KneeXray & 19.25 & 19.09 & 21.61 & 17.78 & 18.95 & \textbf{29.09} \\
Kvasir & 67.34 & 66.50 & 68.88 & 69.65 & 70.81 & 70.86 \\
LungColon & 80.20 & 75.82 & 83.71 & 83.71 & 82.74 & 83.15 \\
OCTMNIST & 46.63 & 47.51 & 49.49 & 52.88 & 49.06 & \textbf{59.43} \\
RETINA & 41.21 & 36.93 & 45.27 & 47.88 & 45.38 & \textbf{54.90} \\
\midrule
\rowcolor{gray!12} \multicolumn{7}{l}{\textbf{Macro-F1: CVaR$_{20}$}} \\
\midrule
BTMRI & 65.89 & 61.39 & 64.61 & 67.41 & 68.38 & \textbf{70.91} \\
BUSI & 47.08 & 48.88 & 49.37 & 50.10 & 48.98 & \textbf{57.94} \\
CHMNIST & 59.94 & 53.27 & 62.48 & 64.20 & 61.85 & 62.11 \\
COVID-19 & 55.21 & 51.65 & 53.00 & 60.84 & 58.83 & \textbf{62.17} \\
CTKidney & 58.64 & 56.50 & 59.67 & 60.99 & 59.17 & \textbf{66.95} \\
DermaMNIST & 21.42 & 19.41 & 25.46 & 25.30 & 22.27 & \textbf{30.03} \\
KneeXray & 16.08 & 16.47 & 18.07 & 17.09 & 15.45 & \textbf{26.35} \\
Kvasir & 64.41 & 64.35 & 65.47 & 66.11 & 67.97 & 67.76 \\
LungColon & 75.64 & 72.50 & 79.39 & 79.09 & 78.25 & 80.06 \\
OCTMNIST & 42.38 & 43.73 & 43.57 & 48.59 & 42.61 & \textbf{54.63} \\
RETINA & 35.24 & 32.98 & 39.55 & 40.63 & 39.15 & \textbf{50.09} \\
\bottomrule
\end{tabular*}
\end{table}

\clearpage
\section{Complete Base-to-Novel Results}
\label{sec:app-base-to-novel}

\begin{center}
\centering
\captionof{table}{Base-to-novel Accuracy for each dataset (\%). Mean and CVaR$_{20}$ are shown in separate blocks; HM is computed from the displayed Base and Novel values. Bold marks the best result in each row.}
\label{tab:app-b2n-accuracy}
\footnotesize
\setlength{\tabcolsep}{0.9pt}
\renewcommand{\arraystretch}{0.76}
\begin{tabular*}{\textwidth}{@{\extracolsep{\fill}}llrrrrrrr>{\columncolor{blue!7}}r@{}}
\toprule
Dataset & Split & BiomedCLIP & CLIP-LoRA & CoOp & CoCoOp & ProGrad & BiomedCoOp & vMFCoOp & \textbf{PARA} \\
\midrule
\rowcolor{gray!12} \multicolumn{10}{l}{\textbf{Mean over support draws}} \\
\midrule
\multirow{3}{*}{BTMRI} & Base & 45.27 & 81.87 & 82.81 & 75.58 & 81.17 & 82.17 & 82.45 & \textbf{86.73} \\
 & Novel & 93.91 & 88.75 & 94.46 & 84.16 & 94.49 & \textbf{95.50} & 94.71 & 93.80 \\
 & \textbf{HM} & 61.10 & 85.17 & 88.25 & 79.64 & 87.33 & 88.34 & 88.16 & \textbf{90.13} \\
\addlinespace[0.5pt]
\multirow{3}{*}{CHMNIST} & Base & 25.51 & 87.01 & 89.84 & 88.99 & 89.08 & 88.55 & 90.89 & \textbf{93.18} \\
 & Novel & 56.98 & 54.34 & 35.54 & 36.73 & 40.98 & 44.68 & 39.98 & \textbf{63.15} \\
 & \textbf{HM} & 35.25 & 66.90 & 50.93 & 52.00 & 56.14 & 59.39 & 55.53 & \textbf{75.28} \\
\addlinespace[0.5pt]
\multirow{3}{*}{COVID-19} & Base & 50.23 & 79.34 & 74.75 & 74.42 & 76.57 & 74.86 & 77.68 & \textbf{81.87} \\
 & Novel & \textbf{91.10} & 70.23 & 89.95 & 89.19 & 89.64 & 90.50 & 91.01 & 90.82 \\
 & \textbf{HM} & 64.75 & 74.51 & 81.65 & 81.13 & 82.59 & 81.94 & 83.82 & \textbf{86.11} \\
\addlinespace[0.5pt]
\multirow{3}{*}{CTKidney} & Base & 56.77 & 85.13 & 80.38 & 81.04 & 76.60 & 84.19 & 86.85 & \textbf{89.43} \\
 & Novel & \textbf{84.36} & 83.99 & 60.79 & 67.46 & 74.53 & 77.24 & 75.38 & 84.20 \\
 & \textbf{HM} & 67.87 & 84.55 & 69.23 & 73.63 & 75.55 & 80.56 & 80.71 & \textbf{86.73} \\
\addlinespace[0.5pt]
\multirow{3}{*}{DermaMNIST} & Base & 57.41 & 56.00 & 49.92 & 40.07 & 52.77 & 57.81 & 51.90 & \textbf{61.98} \\
 & Novel & 65.30 & 58.01 & 70.66 & \textbf{73.39} & 55.64 & 60.69 & 45.23 & 58.93 \\
 & \textbf{HM} & \textbf{61.10} & 56.98 & 58.51 & 51.84 & 54.17 & 59.21 & 48.33 & 60.42 \\
\addlinespace[0.5pt]
\multirow{3}{*}{KneeXray} & Base & \textbf{47.45} & 39.21 & 35.90 & 36.56 & 39.53 & 41.64 & 39.94 & 43.91 \\
 & Novel & 81.94 & 45.78 & 40.09 & 47.14 & 29.28 & 70.12 & 57.93 & \textbf{81.98} \\
 & \textbf{HM} & \textbf{60.10} & 42.24 & 37.88 & 41.18 & 33.64 & 52.25 & 47.28 & 57.19 \\
\addlinespace[0.5pt]
\multirow{3}{*}{Kvasir} & Base & 81.21 & 85.83 & 85.71 & 85.78 & 86.21 & 87.06 & 87.83 & \textbf{88.82} \\
 & Novel & 68.79 & 69.32 & 56.60 & 52.46 & 55.92 & 59.08 & 60.07 & \textbf{71.69} \\
 & \textbf{HM} & 74.49 & 76.70 & 68.18 & 65.10 & 67.84 & 70.39 & 71.34 & \textbf{79.34} \\
\addlinespace[0.5pt]
\multirow{3}{*}{LungColon} & Base & 66.76 & 94.51 & 89.90 & 90.02 & 93.31 & 93.10 & 93.46 & \textbf{96.86} \\
 & Novel & 95.24 & 93.75 & 92.84 & 93.96 & 95.37 & \textbf{97.99} & 97.57 & 95.91 \\
 & \textbf{HM} & 78.50 & 94.13 & 91.35 & 91.95 & 94.33 & 95.48 & 95.47 & \textbf{96.39} \\
\addlinespace[0.5pt]
\multirow{3}{*}{OCTMNIST} & Base & 76.73 & \textbf{89.08} & 67.41 & 74.04 & 68.52 & 74.35 & 76.04 & 87.91 \\
 & Novel & 85.76 & 83.82 & 85.29 & 79.73 & 85.29 & 84.18 & 78.02 & \textbf{85.84} \\
 & \textbf{HM} & 80.99 & 86.37 & 75.30 & 76.78 & 75.99 & 78.96 & 77.01 & \textbf{86.86} \\
\addlinespace[0.5pt]
\multirow{3}{*}{RETINA} & Base & 51.36 & 82.07 & 68.75 & 65.50 & 70.94 & 67.58 & 73.58 & \textbf{83.11} \\
 & Novel & 75.73 & \textbf{77.61} & 59.70 & 59.51 & 62.37 & 67.17 & 70.44 & 76.91 \\
 & \textbf{HM} & 61.21 & 79.78 & 63.90 & 62.36 & 66.38 & 67.38 & 71.98 & \textbf{79.89} \\
\midrule
\rowcolor{gray!12} \multicolumn{10}{l}{\textbf{CVaR$_{20}$ over support draws}} \\
\midrule
\multirow{3}{*}{BTMRI} & Base & 45.27 & 77.45 & 78.74 & 70.02 & 76.15 & 77.47 & 77.48 & \textbf{83.52} \\
 & Novel & 93.91 & 86.12 & 93.76 & 76.59 & 93.38 & \textbf{94.31} & 92.70 & 93.39 \\
 & \textbf{HM} & 61.10 & 81.55 & 85.60 & 73.16 & 83.89 & 85.06 & 84.41 & \textbf{88.18} \\
\addlinespace[0.5pt]
\multirow{3}{*}{CHMNIST} & Base & 25.51 & 84.44 & 86.17 & 85.70 & 85.22 & 86.23 & 87.52 & \textbf{89.89} \\
 & Novel & 56.98 & 51.26 & 31.30 & 27.76 & 35.13 & 39.73 & 34.70 & \textbf{62.69} \\
 & \textbf{HM} & 35.25 & 63.80 & 45.92 & 41.94 & 49.75 & 54.39 & 49.69 & \textbf{73.86} \\
\addlinespace[0.5pt]
\multirow{3}{*}{COVID-19} & Base & 50.23 & 76.09 & 71.67 & 71.84 & 73.61 & 71.40 & 75.81 & \textbf{77.74} \\
 & Novel & \textbf{91.10} & 50.96 & 89.26 & 88.10 & 88.90 & 89.20 & 89.89 & 90.70 \\
 & \textbf{HM} & 64.75 & 61.04 & 79.51 & 79.15 & 80.53 & 79.32 & 82.25 & \textbf{83.72} \\
\addlinespace[0.5pt]
\multirow{3}{*}{CTKidney} & Base & 56.77 & 76.72 & 71.35 & 72.21 & 67.28 & 76.55 & 79.91 & \textbf{80.99} \\
 & Novel & \textbf{84.36} & 83.08 & 43.03 & 58.32 & 62.74 & 69.38 & 67.00 & 84.07 \\
 & \textbf{HM} & 67.87 & 79.77 & 53.69 & 64.52 & 64.93 & 72.79 & 72.89 & \textbf{82.50} \\
\addlinespace[0.5pt]
\multirow{3}{*}{DermaMNIST} & Base & \textbf{57.41} & 52.39 & 44.70 & 33.23 & 48.39 & 52.71 & 45.19 & 57.22 \\
 & Novel & \textbf{65.30} & 50.67 & 57.23 & 61.73 & 22.99 & 48.06 & 28.29 & 58.04 \\
 & \textbf{HM} & \textbf{61.10} & 51.51 & 50.19 & 43.21 & 31.17 & 50.28 & 34.80 & 57.63 \\
\addlinespace[0.5pt]
\multirow{3}{*}{KneeXray} & Base & \textbf{47.45} & 35.84 & 33.70 & 33.01 & 35.74 & 38.64 & 35.37 & 40.06 \\
 & Novel & \textbf{81.94} & 34.32 & 22.33 & 35.16 & 19.00 & 60.53 & 34.92 & 81.92 \\
 & \textbf{HM} & \textbf{60.10} & 35.07 & 26.86 & 34.05 & 24.81 & 47.17 & 35.14 & 53.80 \\
\addlinespace[0.5pt]
\multirow{3}{*}{Kvasir} & Base & 81.21 & 84.80 & 83.79 & 84.07 & 84.20 & 85.39 & 86.24 & \textbf{86.98} \\
 & Novel & 68.79 & 67.32 & 52.20 & 48.32 & 52.98 & 54.85 & 55.98 & \textbf{70.96} \\
 & \textbf{HM} & 74.49 & 75.05 & 64.32 & 61.36 & 65.04 & 66.80 & 67.89 & \textbf{78.16} \\
\addlinespace[0.5pt]
\multirow{3}{*}{LungColon} & Base & 66.76 & 93.53 & 87.86 & 88.40 & 91.64 & 91.65 & 92.14 & \textbf{95.83} \\
 & Novel & 95.24 & 89.09 & 89.48 & 85.76 & 92.33 & \textbf{95.90} & 95.88 & 95.62 \\
 & \textbf{HM} & 78.50 & 91.26 & 88.66 & 87.06 & 91.99 & 93.72 & 93.97 & \textbf{95.73} \\
\addlinespace[0.5pt]
\multirow{3}{*}{OCTMNIST} & Base & 76.73 & \textbf{85.89} & 61.02 & 68.41 & 61.67 & 69.47 & 69.48 & 82.98 \\
 & Novel & 85.76 & 75.57 & 85.29 & 69.37 & 85.29 & 80.48 & 66.54 & \textbf{85.79} \\
 & \textbf{HM} & 80.99 & 80.40 & 71.14 & 68.89 & 71.58 & 74.57 & 67.98 & \textbf{84.36} \\
\addlinespace[0.5pt]
\multirow{3}{*}{RETINA} & Base & 51.36 & 79.05 & 64.77 & 61.31 & 65.58 & 63.99 & 70.49 & \textbf{79.47} \\
 & Novel & 75.73 & 75.59 & 57.14 & 54.07 & 60.07 & 63.09 & 62.35 & \textbf{76.42} \\
 & \textbf{HM} & 61.21 & 77.28 & 60.72 & 57.46 & 62.70 & 63.53 & 66.17 & \textbf{77.91} \\
\bottomrule
\end{tabular*}
\end{center}

\clearpage
\begin{center}
\centering
\captionof{table}{Base-to-novel Macro-F1 for each dataset (\%). Mean and CVaR$_{20}$ are shown in separate blocks; HM is computed from the displayed Base and Novel values. Bold marks the best result in each row.}
\label{tab:app-b2n-macro-f1}
\footnotesize
\setlength{\tabcolsep}{0.9pt}
\renewcommand{\arraystretch}{0.76}
\begin{tabular*}{\textwidth}{@{\extracolsep{\fill}}llrrrrrrr>{\columncolor{blue!7}}r@{}}
\toprule
Dataset & Split & BiomedCLIP & CLIP-LoRA & CoOp & CoCoOp & ProGrad & BiomedCoOp & vMFCoOp & \textbf{PARA} \\
\midrule
\rowcolor{gray!12} \multicolumn{10}{l}{\textbf{Mean over support draws}} \\
\midrule
\multirow{3}{*}{BTMRI} & Base & 45.27 & 81.79 & 82.68 & 75.00 & 80.82 & 82.04 & 82.24 & \textbf{86.71} \\
 & Novel & 93.89 & 88.53 & 94.45 & 82.43 & 94.49 & \textbf{95.48} & 94.70 & 93.77 \\
 & \textbf{HM} & 61.09 & 85.03 & 88.17 & 78.54 & 87.12 & 88.25 & 88.03 & \textbf{90.10} \\
\addlinespace[0.5pt]
\multirow{3}{*}{CHMNIST} & Base & 20.01 & 86.91 & 89.75 & 88.72 & 88.96 & 88.50 & 90.67 & \textbf{93.18} \\
 & Novel & 57.24 & 51.46 & 30.07 & 31.56 & 36.62 & 40.24 & 34.49 & \textbf{62.96} \\
 & \textbf{HM} & 29.65 & 64.64 & 45.05 & 46.56 & 51.88 & 55.32 & 49.97 & \textbf{75.15} \\
\addlinespace[0.5pt]
\multirow{3}{*}{COVID-19} & Base & 44.34 & 78.35 & 71.86 & 72.11 & 74.06 & 70.65 & 76.08 & \textbf{80.84} \\
 & Novel & 75.21 & 60.19 & 59.92 & 55.99 & 58.26 & 64.84 & 70.10 & \textbf{77.58} \\
 & \textbf{HM} & 55.79 & 68.08 & 65.35 & 63.04 & 65.22 & 67.62 & 72.97 & \textbf{79.18} \\
\addlinespace[0.5pt]
\multirow{3}{*}{CTKidney} & Base & 56.68 & 82.88 & 77.13 & 78.32 & 74.96 & 81.79 & 84.06 & \textbf{87.53} \\
 & Novel & 80.28 & \textbf{81.22} & 59.78 & 64.76 & 71.97 & 74.36 & 72.90 & 80.70 \\
 & \textbf{HM} & 66.44 & 82.04 & 67.35 & 70.90 & 73.43 & 77.90 & 78.08 & \textbf{83.98} \\
\addlinespace[0.5pt]
\multirow{3}{*}{DermaMNIST} & Base & 29.98 & 48.25 & 46.12 & 38.23 & 48.62 & 50.32 & 46.39 & \textbf{53.83} \\
 & Novel & 37.25 & 34.03 & \textbf{42.33} & 33.89 & 35.52 & 33.03 & 27.49 & 35.03 \\
 & \textbf{HM} & 33.22 & 39.91 & \textbf{44.15} & 35.93 & 41.05 & 39.88 & 34.52 & 42.44 \\
\addlinespace[0.5pt]
\multirow{3}{*}{KneeXray} & Base & 34.51 & 37.82 & 33.05 & 32.53 & 32.73 & 31.38 & 31.86 & \textbf{41.62} \\
 & Novel & \textbf{54.49} & 43.23 & 33.88 & 36.78 & 26.25 & 51.34 & 45.71 & 53.11 \\
 & \textbf{HM} & 42.26 & 40.34 & 33.46 & 34.53 & 29.14 & 38.95 & 37.55 & \textbf{46.67} \\
\addlinespace[0.5pt]
\multirow{3}{*}{Kvasir} & Base & 80.85 & 85.68 & 85.61 & 85.65 & 86.09 & 86.91 & 87.76 & \textbf{88.69} \\
 & Novel & 68.56 & 69.36 & 49.67 & 46.81 & 49.48 & 54.32 & 57.75 & \textbf{71.54} \\
 & \textbf{HM} & 74.20 & 76.66 & 62.86 & 60.54 & 62.84 & 66.86 & 69.66 & \textbf{79.20} \\
\addlinespace[0.5pt]
\multirow{3}{*}{LungColon} & Base & 66.83 & 94.51 & 89.78 & 89.89 & 93.31 & 93.05 & 93.41 & \textbf{96.86} \\
 & Novel & 95.23 & 93.71 & 92.80 & 93.64 & 95.35 & \textbf{97.98} & 97.57 & 95.91 \\
 & \textbf{HM} & 78.54 & 94.11 & 91.26 & 91.72 & 94.32 & 95.45 & 95.45 & \textbf{96.38} \\
\addlinespace[0.5pt]
\multirow{3}{*}{OCTMNIST} & Base & 64.93 & \textbf{86.00} & 65.19 & 70.32 & 66.35 & 71.06 & 72.17 & 84.77 \\
 & Novel & 51.31 & \textbf{65.94} & 46.03 & 47.27 & 46.03 & 46.95 & 50.48 & 52.28 \\
 & \textbf{HM} & 57.32 & \textbf{74.65} & 53.96 & 56.54 & 54.35 & 56.54 & 59.41 & 64.67 \\
\addlinespace[0.5pt]
\multirow{3}{*}{RETINA} & Base & 47.55 & 81.98 & 68.21 & 64.67 & 70.70 & 66.23 & 73.34 & \textbf{83.04} \\
 & Novel & 74.27 & \textbf{76.62} & 51.10 & 51.70 & 56.28 & 63.10 & 67.32 & 75.67 \\
 & \textbf{HM} & 57.98 & \textbf{79.21} & 58.43 & 57.46 & 62.67 & 64.63 & 70.20 & 79.18 \\
\midrule
\rowcolor{gray!12} \multicolumn{10}{l}{\textbf{CVaR$_{20}$ over support draws}} \\
\midrule
\multirow{3}{*}{BTMRI} & Base & 45.27 & 77.27 & 78.39 & 68.56 & 75.13 & 77.14 & 76.91 & \textbf{83.48} \\
 & Novel & 93.89 & 85.75 & 93.76 & 71.82 & 93.37 & \textbf{94.27} & 92.66 & 93.35 \\
 & \textbf{HM} & 61.09 & 81.29 & 85.39 & 70.15 & 83.27 & 84.85 & 84.05 & \textbf{88.14} \\
\addlinespace[0.5pt]
\multirow{3}{*}{CHMNIST} & Base & 20.01 & 84.17 & 85.97 & 85.25 & 85.00 & 86.05 & 86.88 & \textbf{89.87} \\
 & Novel & 57.24 & 46.96 & 25.18 & 22.10 & 30.07 & 35.10 & 27.84 & \textbf{62.52} \\
 & \textbf{HM} & 29.65 & 60.29 & 38.95 & 35.10 & 44.42 & 49.86 & 42.17 & \textbf{73.74} \\
\addlinespace[0.5pt]
\multirow{3}{*}{COVID-19} & Base & 44.34 & 74.56 & 67.75 & 68.87 & 69.87 & 64.86 & 73.97 & \textbf{76.25} \\
 & Novel & 75.21 & 46.47 & 54.66 & 51.17 & 52.15 & 56.05 & 62.72 & \textbf{77.37} \\
 & \textbf{HM} & 55.79 & 57.26 & 60.51 & 58.71 & 59.72 & 60.14 & 67.89 & \textbf{76.80} \\
\addlinespace[0.5pt]
\multirow{3}{*}{CTKidney} & Base & 56.68 & 74.96 & 68.96 & 70.33 & 66.47 & 74.74 & 77.51 & \textbf{79.17} \\
 & Novel & 80.28 & 80.29 & 41.02 & 55.92 & 62.35 & 67.43 & 65.41 & \textbf{80.57} \\
 & \textbf{HM} & 66.44 & 77.54 & 51.44 & 62.30 & 64.34 & 70.90 & 70.94 & \textbf{79.87} \\
\addlinespace[0.5pt]
\multirow{3}{*}{DermaMNIST} & Base & 29.98 & 45.24 & 42.70 & 33.15 & 45.79 & 47.55 & 42.92 & \textbf{50.13} \\
 & Novel & \textbf{37.25} & 31.32 & 36.88 & 30.14 & 21.95 & 27.53 & 20.62 & 34.65 \\
 & \textbf{HM} & 33.22 & 37.02 & 39.58 & 31.57 & 29.67 & 34.87 & 27.86 & \textbf{40.98} \\
\addlinespace[0.5pt]
\multirow{3}{*}{KneeXray} & Base & 34.51 & 35.21 & 31.16 & 30.10 & 28.47 & 29.32 & 27.41 & \textbf{37.93} \\
 & Novel & \textbf{54.49} & 34.02 & 20.26 & 29.70 & 16.11 & 44.65 & 29.80 & 52.78 \\
 & \textbf{HM} & 42.26 & 34.60 & 24.55 & 29.90 & 20.58 & 35.39 & 28.55 & \textbf{44.14} \\
\addlinespace[0.5pt]
\multirow{3}{*}{Kvasir} & Base & 80.85 & 84.58 & 83.70 & 83.85 & 84.07 & 85.06 & 86.18 & \textbf{86.81} \\
 & Novel & 68.56 & 67.26 & 43.75 & 42.55 & 45.92 & 49.52 & 52.82 & \textbf{70.87} \\
 & \textbf{HM} & 74.20 & 74.94 & 57.46 & 56.45 & 59.40 & 62.60 & 65.50 & \textbf{78.03} \\
\addlinespace[0.5pt]
\multirow{3}{*}{LungColon} & Base & 66.83 & 93.54 & 87.65 & 88.19 & 91.68 & 91.56 & 92.07 & \textbf{95.84} \\
 & Novel & 95.23 & 88.96 & 89.38 & 84.22 & 92.29 & \textbf{95.87} & 95.87 & 95.62 \\
 & \textbf{HM} & 78.54 & 91.19 & 88.51 & 86.16 & 91.98 & 93.67 & 93.93 & \textbf{95.73} \\
\addlinespace[0.5pt]
\multirow{3}{*}{OCTMNIST} & Base & 64.93 & \textbf{82.61} & 59.82 & 65.99 & 60.48 & 67.15 & 66.81 & 79.54 \\
 & Novel & 51.31 & \textbf{59.57} & 46.03 & 43.49 & 46.03 & 46.04 & 45.81 & 51.78 \\
 & \textbf{HM} & 57.32 & \textbf{69.23} & 52.03 & 52.43 & 52.27 & 54.62 & 54.35 & 62.72 \\
\addlinespace[0.5pt]
\multirow{3}{*}{RETINA} & Base & 47.55 & 78.77 & 63.67 & 59.86 & 65.08 & 61.47 & 70.07 & \textbf{79.24} \\
 & Novel & 74.27 & 74.19 & 46.54 & 43.60 & 53.13 & 57.23 & 55.46 & \textbf{75.09} \\
 & \textbf{HM} & 57.98 & 76.41 & 53.78 & 50.45 & 58.51 & 59.28 & 61.91 & \textbf{77.11} \\
\bottomrule
\end{tabular*}
\end{center}

\end{document}